\documentclass[sigconf]{acmart}
\AtBeginDocument{%
  }

\usepackage{graphicx}
\usepackage{makecell}
\usepackage{multirow}
\usepackage{booktabs}
\usepackage{marvosym}
\usepackage{xcolor}
\usepackage{colortbl}
\usepackage{array}
\usepackage{siunitx}
\usepackage{lipsum}
\usepackage{bm}
\usepackage{amsmath}
\usepackage{amsfonts}
\RequirePackage{makecell}
\usepackage{xcolor,colortbl}

\usepackage{bbding}
\usepackage{color}
\usepackage{multicol}

\usepackage{amssymb}
\usepackage{fontawesome}
\usepackage{academicons}

\begin{document}


\title{FastRSR: Efficient and Accurate Road Surface Reconstruction from Bird’s Eye View}


\author{Yuting Zhao$^{*}$}
\affiliation{%
  \institution{Institute of Automation, CAS\\
  School of Artificial Intelligence, UCAS}
  \country{}}
\email{zhaoyuting2023@ia.ac.cn}
  
\author{Yuheng Ji$^{*}$}
\affiliation{%
  \institution{Institute of Automation, CAS\\
  School of Artificial Intelligence, UCAS}
  \country{}
  }
\email{jiyuheng2023@ia.ac.cn}
  
\author{Xiaoshuai Hao}
\affiliation{%
  \institution{Beijing Academy of Artificial Intelligence}
  \country{}}
\email{xshao@baai.ac.cn}

\author{Shuxiao Li$^{\dagger}$}
\affiliation{%
  \institution{Institute of Automation, CAS}
  \country{}}
\email{shuxiao.li@ia.ac.cn}

\renewcommand{\shortauthors}{Zhao et al.}

\begin{abstract}

Road Surface Reconstruction (RSR) is crucial for autonomous driving, enabling the understanding of road surface conditions. 
Recently, RSR from the Bird's Eye View (BEV) has gained attention for its potential to enhance performance. 
However, existing methods for transforming perspective views to BEV face challenges such as information loss and representation sparsity.
Moreover, stereo matching in BEV is limited by the need to balance accuracy with inference speed.
To address these challenges, we propose two efficient and accurate BEV-based RSR models: \textbf{\textit{FastRSR-mono}} and \textit{\textbf{FastRSR-stereo}}.
Specifically, we first introduce Depth-Aware Projection (DAP), an efficient view transformation strategy designed to mitigate information loss and sparsity by querying depth and image features to aggregate BEV data within specific road surface regions using a pre-computed look-up table.
To optimize accuracy and speed in stereo matching, we design the Spatial Attention Enhancement (SAE) and Confidence Attention Generation (CAG) modules. 
SAE adaptively highlights important regions, while CAG focuses on high-confidence predictions and filters out irrelevant information.
FastRSR achieves state-of-the-art performance, exceeding monocular competitors by over \textbf{\textit{6.0\%}} in elevation absolute error and providing at least a \textit{\textbf{3.0}}$\boldsymbol{\times}$ speedup by stereo methods on the RSRD dataset. The source code will be released.

\let\thefootnote\relax\footnotetext{$^{*}$ Equal Contribution.}
\let\thefootnote\relax\footnotetext{$^{\dagger}$ Corresponding Author.}

\end{abstract}



\maketitle

\section{Introduction}

Road Surface Reconstruction (RSR) provides valuable and precise information about road surface conditions, which is essential for planning and control in autonomous driving systems~\cite{oniga2009processing,hao2024your,roadbev}. Recently, real-time reconstruction of road surface geometric elevation information has gained increasing attention due to its role in enhancing driving safety and comfort~\cite{beketov2023impact,alrajhi2023detection,hao2024mapdistill}. Therefore, it is crucial to balance the efficiency and accuracy of road surface reconstruction models for practical deployment.

\begin{figure}[!t]
  \centering
  \includegraphics[width=\linewidth]{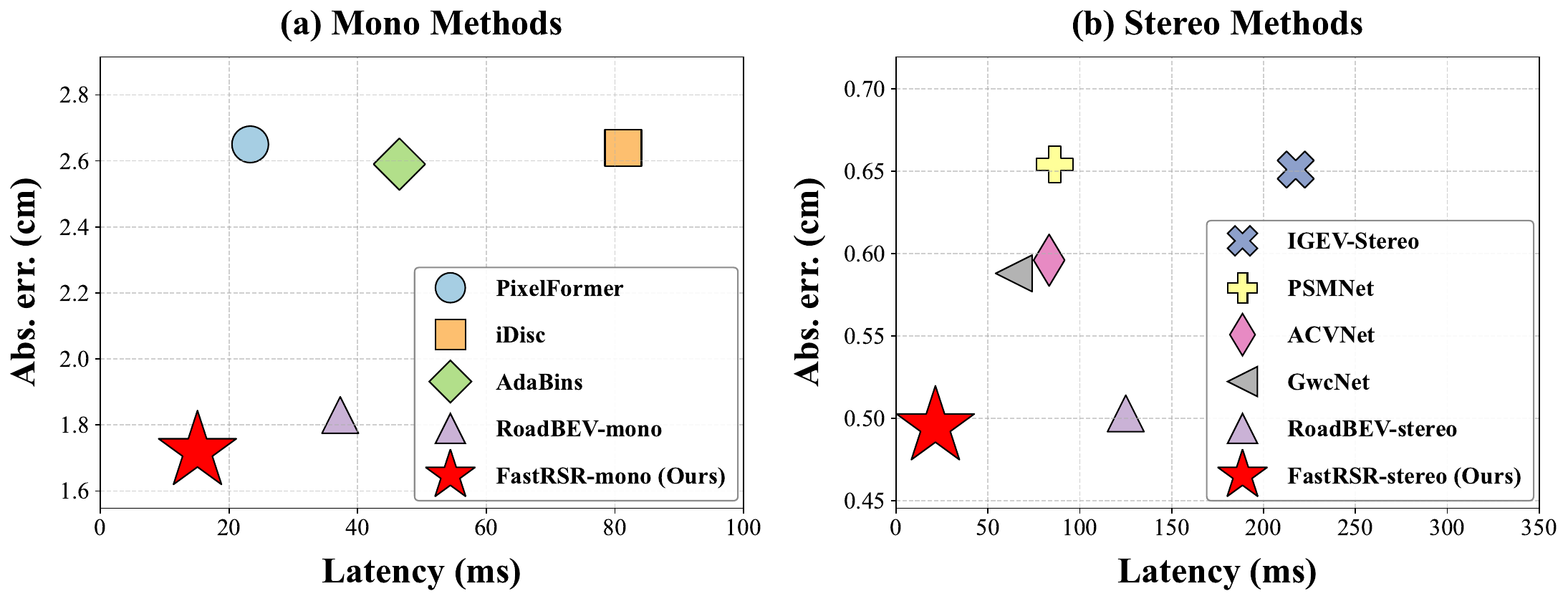}
\caption{Accuracy vs. Latency on the RSRD~\cite{rsrd}. {\bf Left}: FastRSR-mono outperforms other state-of-the-art monocular depth estimation methods. {\bf Right}: FastRSR-stereo surpasses published state-of-the-art stereo matching methods.}
    \label{fig:1}
 \vspace{-1.2em}
\end{figure}

Vision-based Road Surface Reconstruction (RSR) methods offer a cost-effective alternative to LiDAR-based approaches~\cite{rsrlidar1,rsrlidar2,rsrlidar3,rsrlidar4}, yet they often rely on predicting depth from the Perspective View (PV) before converting it to elevation. This process typically results in sparse informative clues for road surface elevation variation~\cite{depth-ele-1,rsrcam2,depth-ele-3}. Recently, RSR from the Bird's Eye View (BEV) has gained attention for its ability to leverage dense elevation cues and accurately capture profile vibrations~\cite{roadbev}.
To facilitate the transformation from PV to BEV, existing View Transformation (VT) methods are categorized into three types: 3D-to-2D projection~\cite{OFTNet,M2BEV,bevformer,simplebev}, 2D-to-3D projection~\cite{lss,bevdet,bevdepth,matrixvt,hao2024mbfusion,bevstereo}, and fusion projection methods~\cite{HAO2025103018,fbbev,dualbev}.
3D-to-2D projection methods use 3D voxel coordinates to query images for dense BEV features; however, the lack of depth guidance can lead to indistinguishable voxel features at varying elevations along the same ray, resulting in information loss. On the other hand, 2D-to-3D projection methods project 2D images into 3D space based on depth distribution, but they often suffer from poor elevation information and representation sparsity. Fusion projection methods attempt to combine features from both projection types, facing challenges in balancing efficiency and accuracy.
This paper explores the question: \textit{Can we design an efficient VT strategy specifically tailored for road surface elevation reconstruction?}

In stereo estimation within the Bird's Eye View (BEV) space, we derive the elevation value for a BEV grid by identifying the most similar matching points among elevation candidates from the left and right voxel features along the vertical dimension. This process parallels PV stereo matching~\cite{rsrcam1,markov,pixelwise}, where the goal is to find corresponding pixels between image pairs that demonstrate the highest similarity across various disparities. Both approaches fundamentally rely on the same principle: searching for the most similar matching points within a defined search range, allowing us to apply concepts from general stereo matching.
However, many existing deep learning-based stereo matching methods typically employ stacked deep 3D convolutional networks to extract effective similarity information from the cost volume, which often results in significant inference time and memory consumption~\cite{deeppruner,psmnet,acvnet}. While there are efficiency-oriented methods that focus on lightweight architectures, these often lead to a considerable degradation in accuracy~\cite{fadnet,mobilestereonet,aanet}. Thus, there is a need for a balance between efficiency and accuracy in stereo estimation for BEV applications.

To address these challenges, we propose two efficient and accurate BEV-based RSR models: \textit{\textbf{FastRSR-mono}} and \textit{\textbf{FastRSR-stereo}}. To address information loss and representation sparsity, we introduce the \textit{Depth-aware 3D-to-2D Projection (DAP)} module. This module effectively queries depth and image features to aggregate BEV features using a pre-computed look-up table within designated road surface profile regions.
For stereo estimation, to construct informative and compact cost volume representation while mitigating performance degradation in lightweight aggregation networks, we first introduce Spatial Attention Enhancement (SAE) to adaptively select important information from the cost volume, utilizing contextual voxel features. Next, we propose Confidence Attention Generation (CAG) to highlight high-confidence predictions while suppressing irrelevant information with grid-wise confidence weights.
Extensive experiments on the real-world RSRD dataset demonstrate the effectiveness of FastRSR, surpassing existing monocular competitors by over 6.0\% in elevation absolute error (\textit{Abs. err.}) and achieving at least a 3.0$\times$ speedup by stereo methods, as shown in Fig.~\ref{fig:1}.

Our main contributions can be summarized as follows:
\begin{itemize}

\item We present two BEV-based RSR models: \textit{\textbf{FastRSR-mono}} and \textit{\textbf{FastRSR-stereo}} for road surface reconstruction, achieving an impressive balance of accuracy and efficiency.
\item We propose an effective view transformation strategy named Depth-Aware 3D-to-2D Projection (DAP), which generates dense BEV features with strong representation ability while enjoying cost-effective deployment.
\item We introduce the Spatial Attention Enhancement (SAE) module to adaptively enhance important spatial regions from the cost volume. Moreover, we propose a Confidence Attention Generation (CAG) module, which emphasizes high-confidence predictions and suppresses irrelevant information to improve feature representation.
\item Our proposed FastRSR achieves superior performance than state-of-the-art (SOTA) methods, which could serve as a strong baseline for road surface reconstruction research.

\end{itemize}

\section{Related Work}
\begin{figure*}[t]
  \centering
  \includegraphics[width=\linewidth]{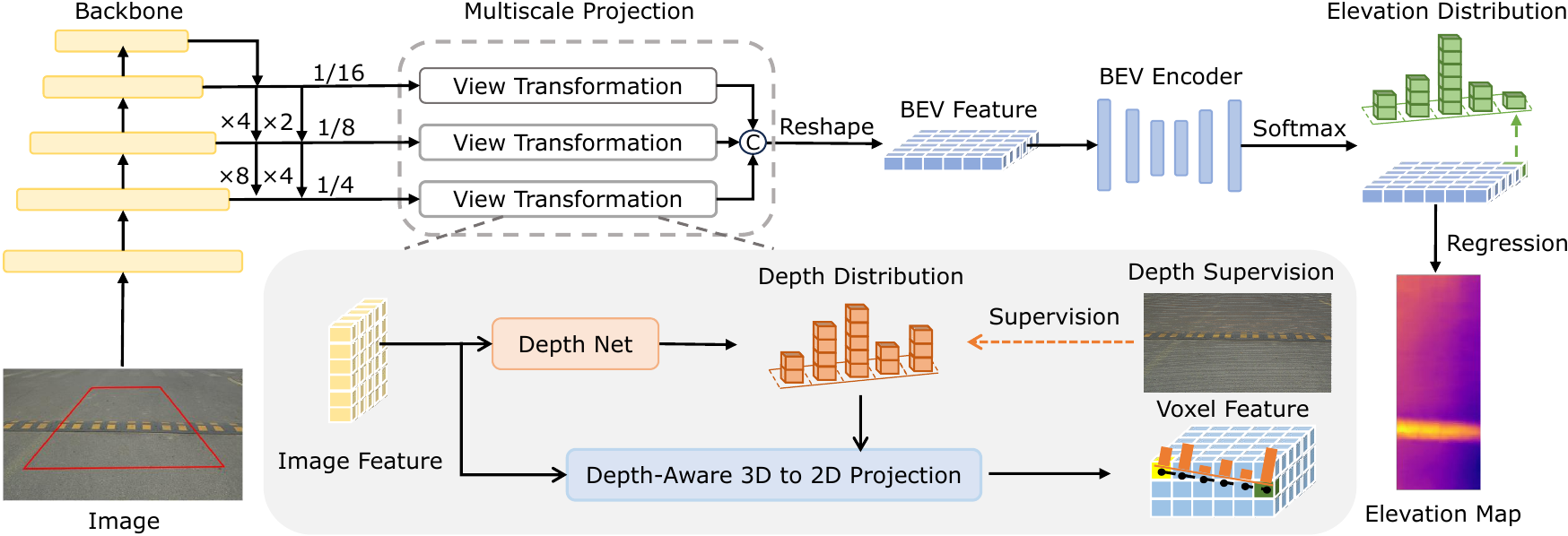}
    \caption{Overview of FastRSR-mono. Based on multi-scale image features, we employ Depth-Aware 3D-to-2D Projection (DAP) for multi-scale projection to extract the rich BEV feature. We then use the BEV encoder and a Softmax function to obtain the elevation distribution. Finally, the predicted elevation map is generated by calculating a linear combination of probability scores utilizing shuttle-shaped discretization bins.}
    \label{fig:mono}
    \vspace{-0.5em}
\end{figure*}

{\bf Vision-Based Road Surface Reconstruction.}
Early research in Road Surface Reconstruction (RSR) introduced geometric prior constraints, utilizing the v-disparity model and visual odometry to extract elevation information from stereo images. However, these methods struggle in textureless scenes with sparse features and suffer from high computational complexity~\cite{rsrcam1,rsrcam2,rsrcam3,pixelwise}.
Recent advancements based on Neural Radiance Fields (NeRF) have optimized 3D road surface rendering using implicit representations and voxel rendering~\cite{nerf,nerf1,nerf2}. Techniques such as plane regularization via Singular Value Decomposition (SVD) have improved NeRF's structural accuracy~\cite{nerf2}. Nevertheless, these approaches are sensitive to camera parameters and often prioritize texture over precise elevation data.
Rome employs mesh partitioning for implicit road surface representation, facilitating large-scale monocular reconstruction~\cite{rome}. The recent RoadBEV method has shown promise in RSR from a Bird's Eye View (BEV)~\cite{roadbev}. However, there remains significant potential for improving processing time and accuracy.

{\bf BEV View Transformation.}
Existing BEV view transformation methods can be categorized into three main types: 3D-to-2D projection, 2D-to-3D projection, and fusion projection methods. In 3D-to-2D projection methods, OFT-Net~\cite{OFTNet} pioneers the use of predefined voxel projections onto images to derive BEV feature representations. Subsequent works~\cite{bevformer,bevformerv2,hao2025msc} build on this approach by employing cross-attention mechanisms within Transformer architectures, though they often overlook depth information during projection, leading to ambiguity. Conversely, 2D-to-3D methods~\cite{lss,bevdepth,bevfusion,bevstereo,chen2025stvit+}, exemplified by LSS~\cite{lss}, project multi-view image features into BEV space based on depth distribution. BEVDepth~\cite{bevdepth} emphasizes depth supervision, while BEVFusion~\cite{bevfusion} extends the LSS framework into the multi-modal perception domain. Despite these advancements, these methods typically yield relatively sparse BEV features. Recent approaches like FB-BEV~\cite{fbbev} and DualBEV~\cite{dualbev} aim to integrate both paradigms. However, a unified view transformation method that effectively combines the strengths of both approaches while ensuring suitability for RSR and fast inference speeds remains elusive.

{\bf Deep Learning-Based Stereo Matching.}
Stereo matching networks based on deep learning have enhanced disparity estimation accuracy~\cite{psmnet,dispnet,deeppruner,igev,acvnet,lightstereo}. For instance, PSMNet~\cite{psmnet} utilizes spatial pyramid pooling and 3D convolutional neural networks to effectively learn disparity information from stereo images. DispNet~\cite{dispnet} employs a correlation layer to directly measure the similarity between left and right images, constructing a single-channel full correlation volume. GWCNet~\cite{gwcnet} introduces a novel method for creating a 4D cost volume using group-wise correlation. Despite these advancements, these methods often struggle with real-time performance.
To address this challenge, some approaches have integrated structures aimed at real-time applications. DeepPruner~\cite{deeppruner} employs a PatchMatch module to incrementally prune the disparity space for each pixel, thereby reducing computational costs. AANet~\cite{aanet} proposes an intra-scale and cross-scale cost aggregation algorithm as an alternative to traditional 3D convolutions, although this comes with a notable sacrifice in accuracy. Nevertheless, current stereo matching methods still face difficulties in achieving an optimal balance between accuracy and computational efficiency.

\section{FastRSR-mono}
In this section, we present the details of FastRSR-mono. We provide an overview of FastRSR-mono in Sec.~\ref{sec:fastrsr-mono}. Specifically, to address the limitations of existing view transformation modules for road surface elevation reconstruction, we introduce the Depth-aware 3D-to-2D Projection (DAP) module in Sec.~\ref{sec:depth_aware}. Considering the distribution characteristics of road surface elevation, we propose the Shuttle-shape Discretization (SD) strategy in Sec.~\ref{sec:elevation regression}.

\vspace{-1em}
\subsection{Overview of FastRSR-mono} \label{sec:fastrsr-mono}
Our goal is to design a novel framework taking image data as input and reconstruct road surface elevation in BEV space. As illustrated in Fig.~\ref{fig:mono}, the monocular image $I\in\mathbb{R}^{H\times W\times 3}$ is fed into the image backbone MobileNetV2~\cite{mobilenetv2} to obtain multi-scale image features $F_{img}=\{\mathbb{R}^{\frac{H}{i}\times\frac{W}{i}\times C_i}|i\in[4,8,16]\}$. Then we obtain the predicted depth distributions $D_{pre}=\{\mathbb{R}^{\frac{H}{i}\times\frac{W}{i}\times C_d}|i\in[4,8,16]\}$ by feeding image features into depth net similar to BEVDepth~\cite{bevdepth}.  To ensure the accuracy of depth distribution estimation, we supervise the intermediate depth prediction $D_{pre}$ using the ground truth depth $D_{gt}$. Taking the image feature $F_{img}$ and depth distribution $D_{pre}$ as input, the DAP module will efficiently generate the strong voxel feature $F_{voxel}\in\mathbb{R}^{N_{x}\times N_{y}\times N_{z} \times C_i}$, as described in Sec.~\ref{sec:depth_aware}. Subsequently, we project multi-scale image features via the DAP module and concact the obtained multi-scale voxel features to form a rich voxel feature representation $\mathbf{B}\in\mathbb{R}^{N_{x}\times N_{y}\times N_{z} \times C}$, where $N_{x}$, $N_{y}$ and $N_{z}$ represent the grid dimensions of the 3D voxel space, and $C$ is three times $ C_i$. We reshape the 4D voxel feature $\mathbf{B}$ into 3D BEV feature $F_{bev}\in\mathbb{R}^{N_{x}\times N_{y} \times (C \times N_{z})}$. We then employ a simplified EfficientNet-B0~\cite{efficientnetb0} as the BEV encoder and a Softmax function to obtain the elevation distribution $E_{prob}\in\mathbb{R}^{N_{x}\times N_{y} \times N}$ over $N$ shuttle-shape discretization bin-centers $e_i, i=1, ..., N$, where $N$ is the number of elevation bins. The elevation bin discretization strategy is detailed in Sec.~\ref{sec:elevation regression}. Finally, the predicted elevation map $E_{pre}\in\mathbb{R}^{N_{x}\times N_{y}}$ is calculated from the linear combination of \textrm{Softmax} scores as follows:
\begin{equation}
E_{pre}(x, y) = \sum_{i=1}^{N} e_i \cdot E_{prob}(x, y, i),
\end{equation}
where $x$ and $y$ represent the lateral and longitudinal coordinates of the BEV grid.

\subsection{Depth-Aware 3D-to-2D Projection} \label{sec:depth_aware}

Given the image features $F_{img}$, we employ a view transformation module to project them into BEV representations. However, current 2D-to-3D projection methods~\cite{lss} tend to produce sparse BEV features, which suffer from poor elevation information. Additionally, existing 3D-to-2D projection methods~\cite{simplebev,bevformer} utilize a bilinear grid sampler or attention operations to query image features for BEV features, which typically takes much time. Moreover, due to the absence of depth guidance, voxel features are assigned the same value at different elevations along the same ray, which leads to information loss and confusion in elevation reconstruction. To address these challenges, we propose a new and powerful Depth-Aware 3D-to-2D Projection (DAP) module to obtain a stronger BEV representation, which is more suitable for road surface elevation reconstruction. Next, we describe the DAP module in detail. 

\begin{figure}[t]
    \centering
    \includegraphics[width=0.98\linewidth]{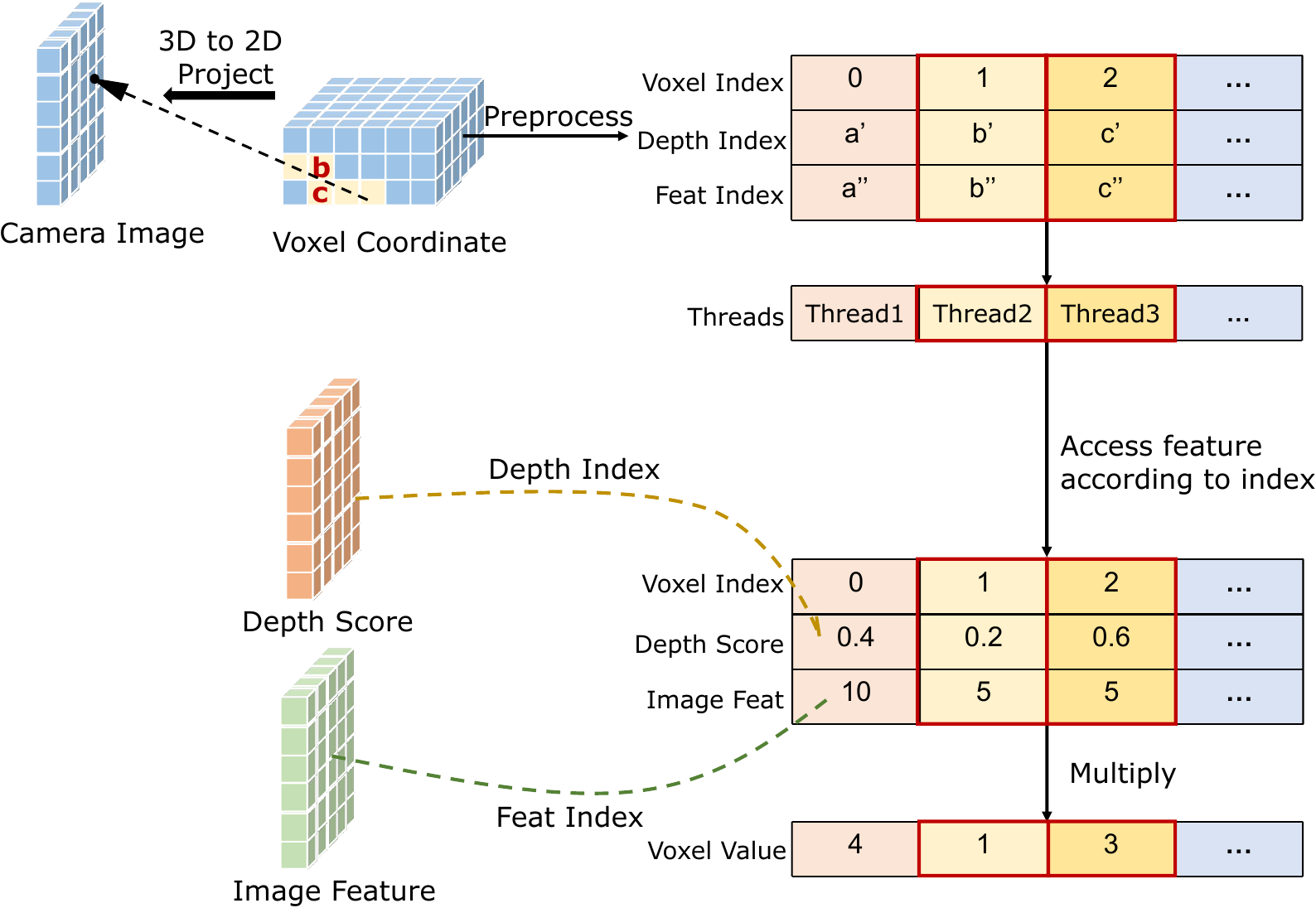}
    \caption{
    Depth-Aware 3D-to-2D Projection Module. Despite occupying different elevations, voxels $b$ and $c$ share identical features along the same ray. With depth information to assign different weights, these voxels receive distinct values.}
    \label{vt}
    \vspace{-0.5em}
\end{figure}

First, to emphasize variations in road surface profiles, we perform voxel partition within a specific road Region of Interest (ROI). Based on the camera intrinsic and extrinsic parameters $K$ and $T$,
we systematically iterate through each voxel cell to calculate the 2D pixel coordinates $p_{2d} = d\cdot(u, v, 1)$ corresponding to the 3D voxel coordinates $p_{3d} = (x, y, z)$ as follows:
\begin{equation}
    d\cdot(u, v, 1)^T=K\cdot T\cdot(x, y, z)^T,
\end{equation}
where $d$ denotes the depth of the 2D pixel point. To distinguish 3D voxels projected onto the same 2D image point along the same ray, we introduce depth information as a key distinguishing factor. Given the image feature $F_{img}$ and depth distribution $D_{pre}$, we employ bilinear grid sampler $S_{2d}(\cdot)$ and trilinear grid sampler $S_{3d}(\cdot)$ to sample the corresponding image feature and depth probability at the 2D pixel point $p_{2d}$. The voxel feature is obtained as follows:   
\begin{equation}
    F_{voxel}(p_{3d}) = S_{2d}(F_{img}, p_{2d}) \cdot S_{3d}(F_d, p_{2d}).
\end{equation}
Since the projection relationship is established without any learnable parameters, we can accelerate the process by pre-computing the fixed projection indexes according to the established 3D-to-2D projection and storing them in a static look-up table. This allows us to substitute the grid sampler operator. Additionally, we use multiple processing threads to further enhance the speed of this procedure. As illustrated in Fig.~\ref{vt}, we retrieve the value of the corresponding depth scores and feature values using the preprocessed image feature index $f^{index}$ and depth index $d^{index}$ within the processing thread. By multiplying these values, we derive the voxel feature values as follows: 
\begin{equation}
    F_{voxel}(p_{3d}) = F_{img}(f^{index}) \cdot F_d(d^{index}).
\end{equation}

\vspace{-1em}
\subsection{Shuttle-shape Discretization}\label{sec:elevation regression}

\begin{figure}[!t]
  \centering
  \includegraphics[width=0.98\linewidth]{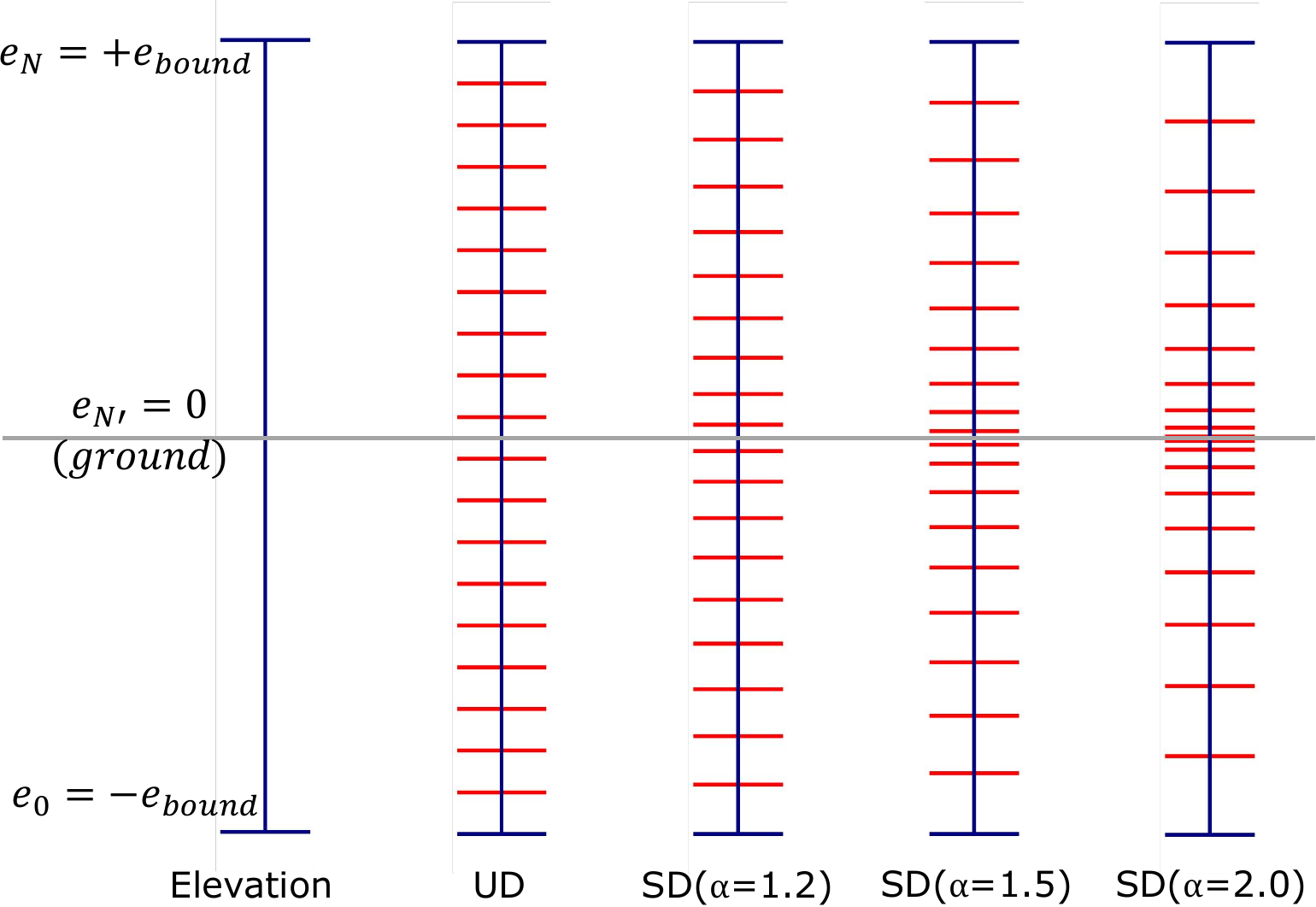}
    \caption{Discretization bin methods.}
    \label{fig:SD}
    \vspace{-0.5em}
\end{figure}

The continuous elevation space is discretized into a set of $N$ bins. Previous elevation discretization is typically performed using Uniform Discretization (UD) with a fixed bin size. However, considering the distribution characteristics of road surface elevation, we propose the Shuttle-shape Discretization (SD) strategy. This approach is centered around a ground elevation value of zero, with bin widths gradually transitioning from dense to sparse towards the upper and lower ends. The elevation discretization techniques are visualized in Fig.~\ref{fig:SD}. SD is defined as follows:
\begin{equation}
b_{i} = 
\begin{cases} 
\left( \frac{N' - N_{i}}{N'} \right)^{\alpha} \cdot e_{bound}, & N_{i} = 0, 1, \ldots, N'-1 \\
-\left( \frac{N_{i} - N'}{N'} \right)^{\alpha} \cdot e_{bound}, & N_{i} = N', N'+1, \ldots, N    
\end{cases}, \nonumber
\end{equation}
\begin{equation}
e_i = \frac{1}{2} \times (b_{i-1} + b_{i}),
\end{equation}
where $b_{i}$ represents the $i^{th}$ elevation bin edge value, $e_{bound}$ denotes the boundary of predefined elevation range, $N' = N/2$ is the half of the number of elevation bins, $N_{i}$ denotes the $i^{th}$ elevation bin, $e_{i}$ is the $i^{th}$ elevation bin center value and $\alpha$ is the hype parameter to control the sensitivity across various elevation ranges, shaping the degree of non-linearity in the output.

\section{FastRSR-stereo}
In this section, for stereo images, we present the details of FastRSR-stereo. We begin with an overview of FastRSR-stereo in Sec.~\ref{sec:fastsr_stereo}. Specifically, to alleviate the performance loss associated with lightweight architectures, we propose the Spatial Attention Enhancement (SAE) module in Sec.~\ref{sec:spatial attention} and the Confidence Attention Generation (CAG) module in Sec.~\ref{sec:attention_generation}.

\subsection{Overview of FastRSR-stereo} \label{sec:fastsr_stereo}

\begin{figure*}[t]
  \centering
  \includegraphics[width=\linewidth]{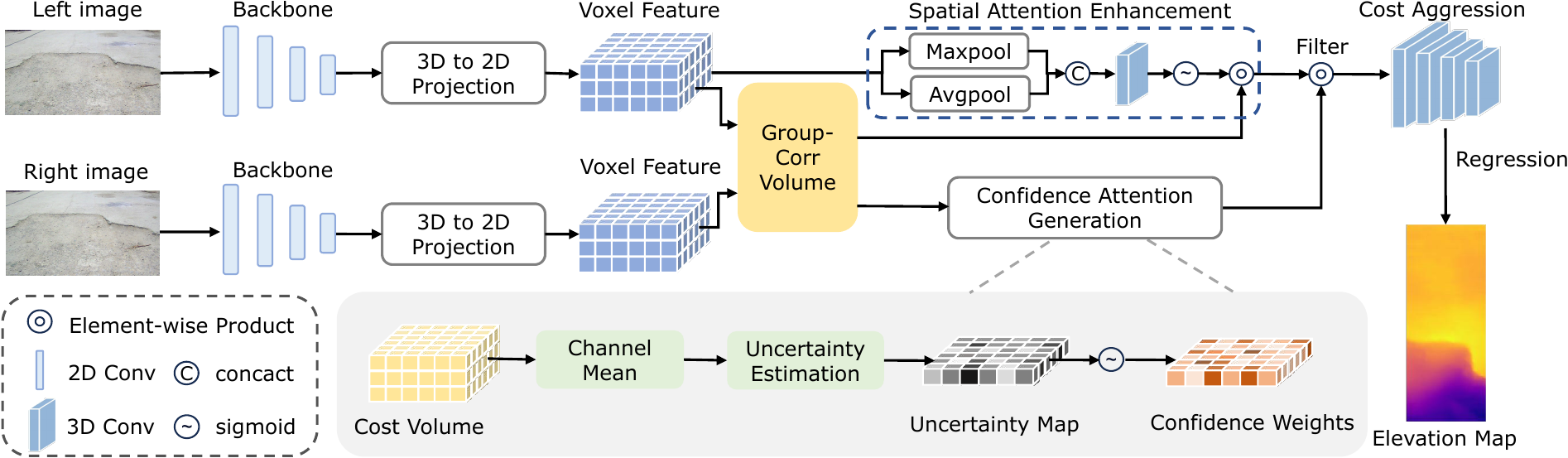}
    \caption{Overview of FastRSR-stereo. Following the paradigm of monocular BEV feature extraction, we first construct the initial group-wise correlation cost volume from the left and right voxel features. Subsequently, we introduce the Spatial Attention Enhancement (SAE) module to adaptively enhance important regions, while the Confidence Attention Generation (CAG) module emphasizes high-confidence predictions and suppresses irrelevant information.}
    \label{fig:stereo}
    \vspace{-0.5em}
\end{figure*}

As illustrated in Fig.~\ref{fig:stereo}, FastRSR-stereo mainly consists of five components: voxel feature extraction, initial volume construction, attention volume construction, cost aggregation and disparity regression. In the following, we provide an overview of each module.

\textbf{Voxel Feature Extraction.}
The stereo branch follows a similar paradigm to monocular BEV feature extraction. For a stereo image pair $I_{l}, I_{r}\in\mathbb{R}^{H\times W\times 3}$, we utilize light backbone MobileNetV2~\cite{mobilenetv2} to obtain image features. Subsequently, we project image features via the Depth-Aware 3D-to-2D
Projection (DAP) module to derive rich voxel features $\mathbf{B}_{l}, \mathbf{B}_{r}\in\mathbb{R}^{N_{x}\times N_{y}\times N_{z} \times C}$.

\textbf{Initial Volume Construction.}
Given the left and right voxel features $\mathbf{B}_{l}$ and $\mathbf{B}_{r}$, we follow the general framework of stereo matching. In the BEV space, the vertical elevation candidate values correspond to disparity values in stereo matching. We construct the initial group-wise correlation cost volume $\mathbf{V}^{init}$~\cite{gwcnet} as follows:
\begin{equation}
    \mathbf{V}^{init}(x, y, z, g) = \frac{1}{G} \left( \mathbf{B}_{l}^g(x, y, z) \odot \mathbf{B}_{r}^g(x, y, z) \right),
\end{equation}
where $G= {N_{c}/N_{g}}$, $N_{c}$ is the number of channels. All the channels are equally divided into $N_g$ groups along the channel dimension. $\odot$ represents the element-wise product from the $g^{th}$ voxel feature group $\mathbf{B}_{l}^g, \mathbf{B}_{r}^g$.

\textbf{Attention Volume Construction.}
We first utilize the Spatial Attention Enhancement (SAE) module to enhance the representation capacity of initial cost volume $\mathbf{V}^{init}$, as detailed in Sec.~\ref{sec:spatial attention}. Subsequently, we implement the Confidence Attention Generation (CAG) module to derive confidence weights and filter the enhanced cost volume for reliable prediction, as described in Sec.~\ref{sec:attention_generation}.

\textbf{Cost Aggregation.}
We employ an efficient cost aggregation network comprising four initial 3D convolutions followed by three lightweight stacked 3D hourglass networks. Each hourglass module is carefully designed with only three 3D convolutions and a single 3D deconvolution operation.

\textbf{Disparity Regression.}
With the aggregated volume, the probability of each elevation disparity is calculated via a Softmax operation. The predicted elevation disparity is derived from the linear combination of Softmax scores.

\subsection{Spatial Attention Enhancement} \label{sec:spatial attention}

To mitigate the performance degradation caused by lightweight backbone and cost aggregation networks, we propose the Spatial Attention Enhancement (SAE) module to achieve more accurate matching information with contextual guidance. Inspired by~\cite{cbam}, we generate spatial attention weights to effectively and adaptively select important regions. Next, we explain our SAE module in detail.

Given the left voxel feature $\mathbf{B}_{l}\in\mathbb{R}^{N_{x}\times N_{y}\times N_{z} \times C}$, we first apply max pooling and average pooling operations along the channel dimension to obtain pooling maps $\mathbf{B}_{avg},\mathbf{B}_{max}\in\mathbb{R}^{N_{x}\times N_{y}\times N_{z} \times 1}$. We then concatenate these maps to form a map in $\mathbb{R}^{N_{x}\times N_{y}\times N_{z} \times 2}$. Subsequently, to derive the spatial attention weights $\mathbf{A^s}$, we perform a convolution operation and a Sigmoid function as follows:
\begin{equation}
    \mathbf{A^s} = \sigma\left(f^{7\times7}\left([Avg(\mathbf{B}_{l}); Max(\mathbf{B}_{l})]\right)\right),
\end{equation}
where $\sigma$ denotes the Sigmoid function and $f^{7\times 7}$ represents a 3D convolution operation with the filter size of $1\times 7 \times 7$. The spatial attention weights encode where to be emphasized or suppressed, allowing for the adaptive selection of important regions. The enhanced cost volume $\mathbf{V^e}$ is computed as follows:
\begin{equation}
    \mathbf{V^e} = \mathbf{A^s} \odot \mathbf{V}^{init}.
\end{equation}

\subsection{Confidence Attention Generation} \label{sec:attention_generation}

Despite the effectiveness of the spatial attention enhancement module, we argue that how to design an effective attention generation module for emphasizing high-confidence predictions while suppressing irrelevant information remains crucial. For stereo estimation in BEV space, we obtain the elevation value of a BEV grid by identifying the most similar matching points among the elevation candidates of left and right voxel features along the vertical dimension. The ideal elevation distribution within a grid should be unimodal, peaking at correctly matched points. Grids with a multimodal distribution are considered unreliable. To emphasize high-confidence predictions, we propose the Confidence Attention Generation (CAG) module.

Given the initial cost volume $\mathbf{V^{init}}\in\mathbb{R}^{N_{g} \times N_{x}\times N_{y} \times N_{z}}$, we first compute the average value across all channels. Then we obtain the elevation probability distribution volume $\mathbf{P}^{init}\in\mathbb{R}^{N_{x}\times N_{y} \times N_{z}}$ by applying the Softmax function. We regress the elevation disparity map $\mathbf{D}^{init}\in\mathbb{R}^{N_{x} \times N_{y}}$ from the linear combination of probabilities $\mathbf{P}^{init}$. The uncertainty is calculated from the variance of the elevation distribution as follows:
\begin{equation}
    \mathbf{U}(i)=\sum_{j=0}^{N_{z}-1}\left(z_j-\mathbf{D}^{init}(i)\right)^2 \times \mathbf {P}_j^{init}(i),
\end{equation}
where $z_j$ is the $j^{th}$  voxel elevation center and ${P}_j^{init}(i)$ represents the $j^{th}$ probability value at the BEV grid $i$. The confidence $\mathbf{A^c}$ at the grid $i$ is defined as follows:
\begin{equation}
    \mathbf{A^c}(i)=\sigma(\epsilon+ s\times\mathbf{U}(i)),
\end{equation}
where $s, \epsilon$ are learnable parameters, and $\sigma$ denotes the Sigmoid function. The confidence attention weights $\mathbf{A^c}\in\mathbb{R}^{ N_x \times N_y}$ indicate that BEV grids with high confidence correspond to reliable identification of a unique match. Finally, the attention volume $\mathbf{V^a}$ is computed as follows:
\begin{equation}
    \mathbf{V^a} = \mathbf{A^c} \odot \mathbf{V}^{e}.
\end{equation}
\section{Training Strategy} \label{sec:loss_function}
Our method adopts an end-to-end training strategy, supervising the intermediate predictions of depth distribution and the final predicted elevation. The ground truth values for depth and elevation are derived from LiDAR point clouds, providing a robust and accurate foundation for training. We employ the cross entropy losses to train the network, our final total loss is defined as:
\begin{equation}
\begin{split}
\mathcal{L} = {} & \frac{1}{N_{grid}}\sum_{i=1}^{N_{grid}}{M}_{e}(i)\cdot \mathrm{CE}({E}_{pre}(i),{E}_{gt}(i)) \\
              & + \beta\cdot\frac{1}{N_{grid}}\sum_{i=1}^{N_{grid}}{M}_{d}(i)\cdot \mathrm{CE}({D}_{pre}(i),{D}_{gt}(i)),
\end{split}
\end{equation}
where $N_{grid} = N_x \times N_y$ is the number of BEV grids, ${E}_{pre}$ denotes the predicted elevation, ${E}_{gt}$ represents the Ground-Truth (GT) elevation value, ${D}_{pre}$ is depth distribution predictions and ${D}_{gt}$ is depth distribution labels. To ensure the accuracy of our predictions, we utilize the binary GT masks ${M}_{e}$ and ${M}_{d}$ to filter valid elevation and depth values. We set $\beta = 0.25$ for all experiments, which balances the contributions of elevation and depth losses in our total loss.

\section{Experiment}

\subsection{Experimental Setup} \label{details}
{\bf Dataset and Metrics.} The RSRD dataset is the large-scale and real-world dataset for road surface reconstruction~\cite{rsrd}. It provides 2,800 pairs of high-resolution stereo images, dense point cloud labels, and motion pose information within a dense subset. For training our network, we utilize a representative subset containing 1,210 training samples and 371 testing samples.
In evaluations, we use the metrics defined as: Absolute Error (Abs. err.): $\frac{1}{N_{grid}}\sum_{i=1}^{N_{grid}}{|{E}_{gt}(i)-{E}_{pre}(i)|}$, Root Mean Squared Error (RMSE): $\sqrt{\frac{1}{N_{grid}}\sum_{i=1}^{N_{grid}}({E}_{gt}(i)-{E}_{pre}(i))^2}$, the percentage of grids with error greater than 0.5 cm (\textgreater0.5 cm), and the GPU latency in inference.

{\bf Implementation Details.} We implement the proposed network operating at a cropped image resolution of 960×528 in PyTorch. For depth-aware 3D-to-2D projection, we define the ROI as $[-1.0 m, 0.9 m]$, $[2.2 m, 7.1 m]$ and $[-0.2 m, 0.2 m]$ with a voxel resolution of $[x_{res}, y_{res}, z_{res}] = [0.03 m, 0.03 m, 0.01 m]$ along $X$, $Y$, $Z$ axes. Our model is trained for 50 epochs using a batch size of 8 across all experiments. We utilize the AdamW optimizer with a weight decay 1e-4, coupled with OneCycleLR scheduling. The maximum learning rate is set to 8e-4 for FastRSR-mono, while 5e-4 for FastRSR-stereo with a linear decreasing strategy. All latency tests are conducted on a single NVIDIA 3090 GPU.

\subsection{Comparisons with State-of-the-art}
We comprehensively compare our proposed models with the baseline method RoadBEV~\cite{roadbev}, several state-of-the-art monocular depth estimation and stereo matching approaches. For a fair comparison, these models follow their original training configurations to obtain depth from the perspective view, which is then converted into BEV to generate the elevation map.

\textbf{Evaluation on Monocular Elevation Estimation.} We compared our FastRSR-mono model with four baseline monocular depth estimation methods using the RSRD dataset in Tab.~\ref{tab:comparison_all}. The results lead us to two key conclusions: (1) \textit{Significant Performance Improvement.} FastRSR-mono achieves a state-of-the-art absolute error (Abs. err.) of 1.72 cm, which surpasses PixelFormer~\cite{pixelformer} by 35.1\%.  This significant improvement highlights the effectiveness of our model in monocular elevation reconstruction.
(2) \textit{Rapid Inference Efficiency.} FastRSR-mono model achieves an inference time of only 15.1 ms, which is substantially lower than other methods, making our monocular model a practical choice for dynamic environments.

\textbf{Evaluation on Stereo Elevation Estimation.} We compared our FastRSR-stereo model with five baseline stereo matching methods using the RSRD dataset in Tab.~\ref{tab:comparison_all}. The results lead us to three key conclusions: (1) \textit{Superior Accuracy Compared with Monocular Methods.} It is evident that stereo matching models generally outperform monocular approaches. Notably, our proposed FastRSR-stereo model exhibits significant advantages, achieving the absolute elevation error of 0.495 cm, the lowest among the evaluated stereo methods. (2) \textit{Real-time Performance.} The inference time for FastRSR-stereo is 21.4 ms, which surpasses existing stereo competitors by at least 3.0${\times}$ speedup. (3) \textit{Limited Accuracy Improvement.} While stereo methods generally outperform monocular ones, the accuracy gains are marginal. A potential reason is that the models already reach very high accuracy, while making further promotion is challenging. Additionally, noise in the labels could hinder further enhancements.

\begin{table}[t]
  \centering\addtolength{\tabcolsep}{-2.0pt}
  \caption{Performance comparison on the RSRD dataset. }
  \scalebox{0.96}{
    \begin{tabular}{c|c|c|c|c|c}
    \hline
    \rowcolor{black!10} \multicolumn{2}{c|}{\textbf{Method}} & \makecell[c]{\textbf{Abs. err.} \\ (\textbf{cm})} & \makecell[c]{\textbf{RMSE} \\ (\textbf{cm})} & \makecell[c]{\textbf{\textgreater0.5 cm} \\ (\textbf{\%})} & \makecell[c]{\textbf{Lantency} \\ (\textbf{ms})}  \\
    \hline
    \multirow{5}{*}{\rotatebox{90}{Mono}}  
     & PixelFormer~\cite{pixelformer} & 2.65 & 2.86 & 82.0 & 23.3\\
     & iDisc~\cite{idisc} & 2.64 & 2.88 & 84.3 & 81.3 \\
     & AdaBins~\cite{adabins} & 2.59 & 2.79 & 82.4 & 46.5 \\
     & RoadBEV-mono~\cite{roadbev} & 1.83 & 2.07 & 78.6 & 37.3 \\
     \rowcolor{blue!10} & \textbf{FastRSR-mono} & \textbf{1.72} & \textbf{1.95} & \textbf{76.3} & \textbf{15.1}  \\
    \hline
    \multirow{6}{*}{\rotatebox{90}{Stereo}} & IGEV-Stereo\cite{igev} & 0.651 & 0.797 & 49.5 & 217.5 \\
     & PSMNet~\cite{psmnet} & 0.654 & 0.785 & 50.1 & 86.2 \\
     & ACVNet~\cite{acvnet} & 0.596 &  0.723 & 46.2 & 83.3 \\
     & GwcNet~\cite{gwcnet} & 0.588 &  0.711 & 44.9 & 64.1 \\
     & RoadBEV-stereo~\cite{roadbev} & 0.503 & 0.609  & 37.0 & 125.0\\
     \rowcolor{blue!10} & \textbf{FastRSR-stereo }& \textbf{0.495} & \textbf{0.605} & \textbf{36.8} & \textbf{21.4}  \\
     
    \hline
    \end{tabular}%
    }
  \label{tab:comparison_all}%
  \vspace{-1em} 
\end{table}%

\subsection{Ablation Studies for FastRSR-mono}

{\bf Contribution of major components in FastRSR-mono.} To verify the effectiveness of our proposed modules, we take RoadBEV-mono~\cite{roadbev} with a light backbone~\cite{mobilenetv2} as our baseline. We conduct an ablation study on the components in FastRSR-mono and summarize our results in Tab.~\ref{ablation-mono}. We evaluate model variants using different combinations of the proposed modules, including Shuttle-shape Discretization (SD), Multi-Scale Projection (MSP) and Depth-Aware 3D-to-2D Projection (DAP).

We first investigate the effect of each component.
In model variants (a), (b), and (c), we add different modules separately.
The experimental results show that all model variants get improved performance compared to the baseline, verifying the effectiveness of the proposed modules.
Moreover, the results of model variants (d), (e), and (f) prove that different modules are complementary to each other.
Finally, using all the proposed modules together, we arrive at the full FastRSR-mono, which achieves the best performance of 1.72 cm, significantly surpassing the baseline of 22.5\%.

\begin{figure}[t]
\centering
\begin{minipage}{0.49\linewidth}
    \centering
    \includegraphics[width=0.9\linewidth]{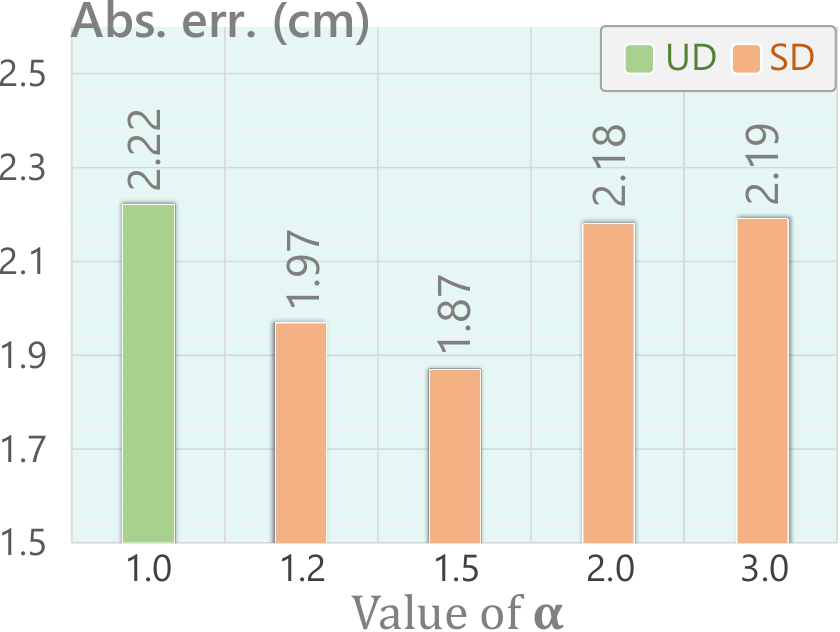}
\end{minipage}
\begin{minipage}{0.49\linewidth}
    \centering
    \includegraphics[width=0.9\linewidth]{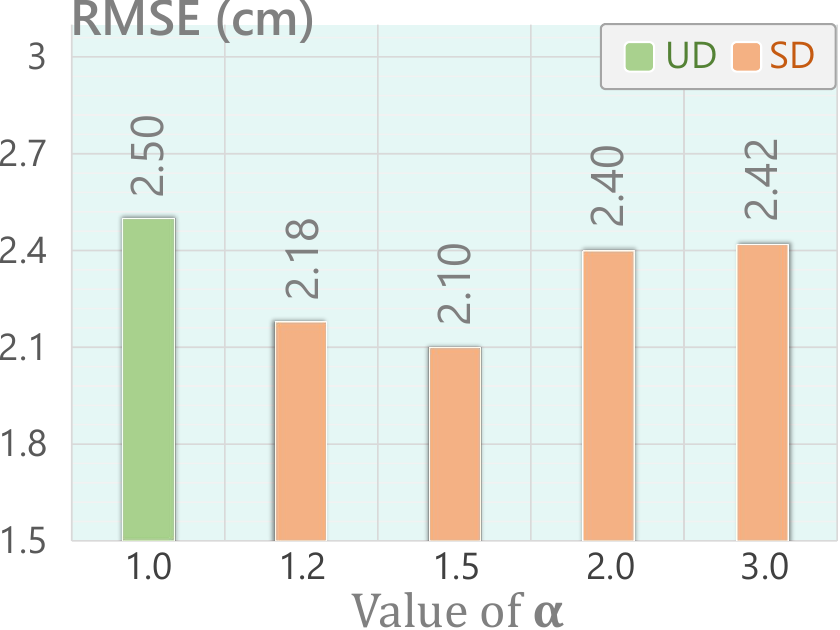}
   
\end{minipage}

\captionsetup{skip=2pt} 
\caption{Comparison of different discretization bin methods.} 
\label{fig:alpha}
\vspace{-0.5em}
\end{figure}

{\bf Discretization Bin Types.}
To evaluate the performance of different discretization bin methods, we perform an ablation study in Fig.\ref{fig:alpha}. The Shuttle-shape Discretization (SD) demonstrates superior adaptability to real-world road geometry compared to Uniform Discretization (UD). When our SD strategy with $\alpha=1.5$ is applied instead of UD, the prediction accuracy improves 15\%. The performance degradation when $\alpha=2.0$ reveals that excessive focus on near-field details introduces noise amplification in distant areas. Ultimately, the performance achieves the best 1.87 cm absolute elevation error and 2.10 cm root mean square error at $\alpha=1.5$.

\begin{table}[t]
\begin{center}
\caption{Ablation study on the components in FastRSR-mono. SD is the shuttle-shape discretization. MSP means using the multi-scale projection. DAP denotes the depth-aware 3D-to-2D projection module.}
\scalebox{0.78}{ 
  \begin{tabular}{p{1.2cm}p{0.6cm}p{0.8cm}p{0.8cm}|p{1.2cm}p{1.1cm}p{1.1cm}p{1.2cm}}
  \hline
   \rowcolor{black!10} \makecell[c]{\textbf{Setting}}&\makecell[c]{\textbf{SD}}& \makecell[c]{\textbf{MSP}}& \makecell[c]{\textbf{DAP}} & \makecell[c]{\textbf{Abs. err.}\\(cm)} &\makecell[c]{\textbf{RMSE}\\(cm)}&\makecell[c]{\textbf{>0.5 cm}\\(\%)}&\makecell[c]{\textbf{Latency}\\(ms)}\\
  \midrule
  
   \makecell[c]{Baseline}&\makecell[c]{\XSolidBrush}& \makecell[c]{\XSolidBrush}& \makecell[c]{\XSolidBrush}& \makecell[c]{2.22} &\makecell[c]{2.50}&\makecell[c]{83.0}&\makecell[c]{14.2}\\
    \hline
\makecell[c]{a}& \makecell[c]{\CheckmarkBold}& \makecell[c]{\XSolidBrush}& \makecell[c]{\XSolidBrush}& \makecell[c]{1.87} &\makecell[c]{2.10}&\makecell[c]{77.5}&\makecell[c]{14.2}\\
   \makecell[c]{b}& \makecell[c]{\XSolidBrush}& \makecell[c]{\CheckmarkBold}& \makecell[c]{\XSolidBrush}& \makecell[c]{1.94} &\makecell[c]{2.16}&\makecell[c]{78.2}&\makecell[c]{14.9}\\
   \makecell[c]{c}&\makecell[c]{\XSolidBrush}& \makecell[c]{\XSolidBrush}& \makecell[c]{\CheckmarkBold}& \makecell[c]{1.91} &\makecell[c]{2.16}&\makecell[c]{78.1}&\makecell[c]{14.2}\\
   \hline
 \makecell[c]{d}&\makecell[c]{\CheckmarkBold}& \makecell[c]{\CheckmarkBold}& \makecell[c]{\XSolidBrush}& \makecell[c]{1.80} &\makecell[c]{2.01}&\makecell[c]{76.4}&\makecell[c]{14.9}\\
   \makecell[c]{e}& \makecell[c]{\XSolidBrush}& \makecell[c]{\CheckmarkBold}& \makecell[c]{\CheckmarkBold}& \makecell[c]{1.82} &\makecell[c]{2.07}&\makecell[c]{77.1}&\makecell[c]{15.1}\\
    \makecell[c]{f}& \makecell[c]{\CheckmarkBold}& \makecell[c]{\XSolidBrush}& \makecell[c]{\CheckmarkBold}& \makecell[c]{1.76} &\makecell[c]{1.98}&\makecell[c]{76.2}&\makecell[c]{14.2}\\
     \hline
   \rowcolor{blue!10}\makecell[c]{g}&\makecell[c]{\CheckmarkBold}& \makecell[c]{\CheckmarkBold}& \makecell[c]{\CheckmarkBold}& \makecell[c]{\textbf{1.72}} &\makecell[c]{\textbf{1.95}}&\makecell[c]{\textbf{76.3}}&\makecell[c]{\textbf{15.1}}\\
  \bottomrule
  \label{ablation-mono}
  \end{tabular}}
\end{center}
 \vspace{-2em}
\end{table}

\begin{figure}[t]
\centering
\begin{minipage}{0.49\linewidth}
    \centering
    \includegraphics[width=0.9\linewidth]{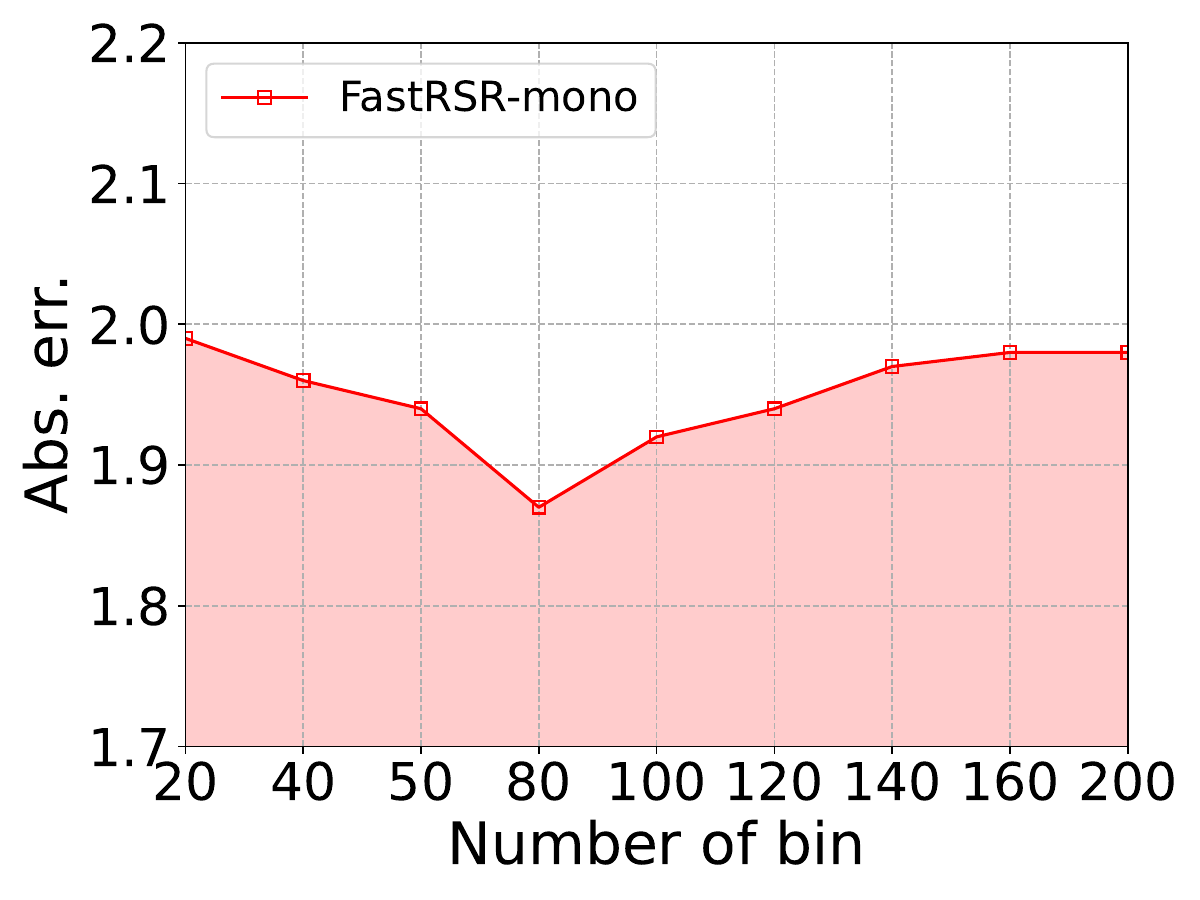}
    \captionsetup{skip=2pt} 
    \caption{Effect of numbins.}
    \label{fig:bin}
\end{minipage}
\begin{minipage}{0.49\linewidth}
    \centering
    \includegraphics[width=0.9\linewidth]{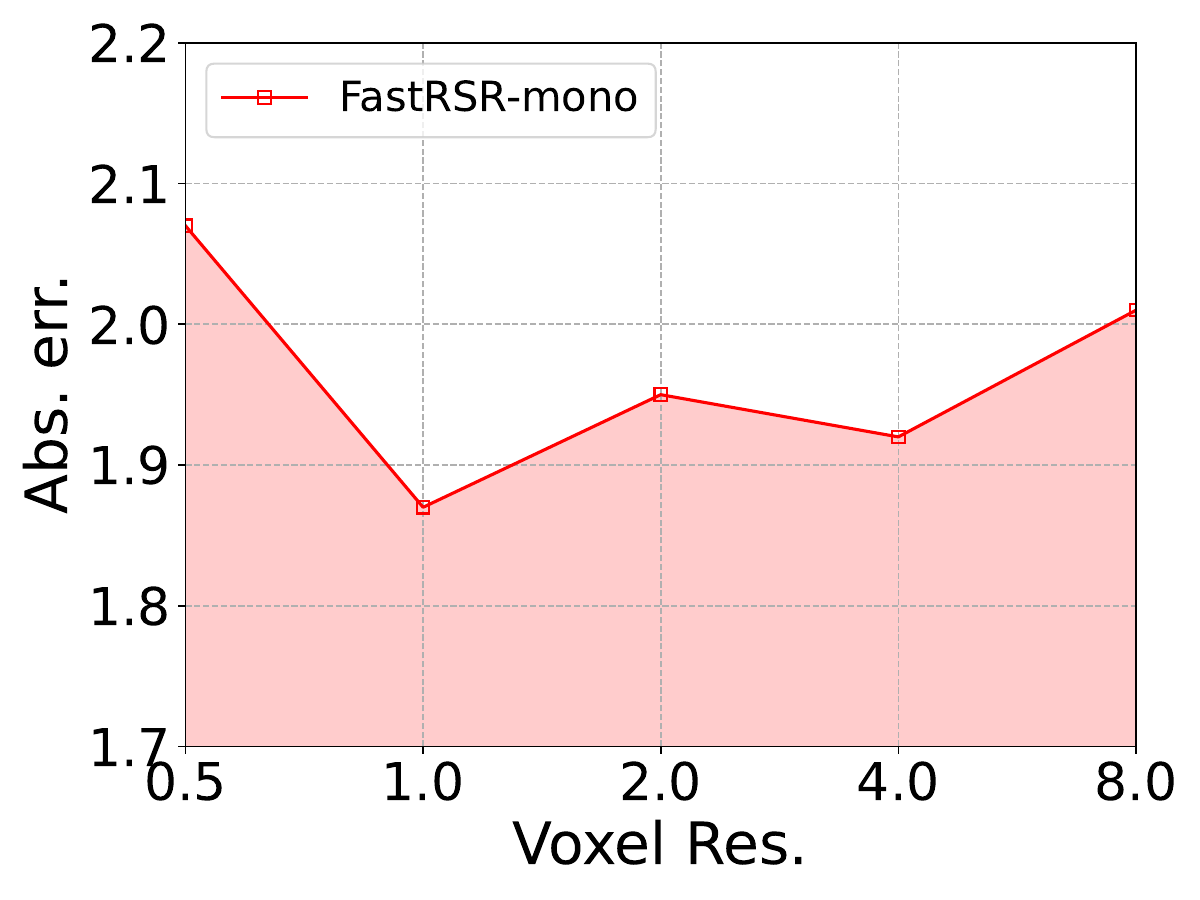}
    \captionsetup{skip=2pt} 
    \caption{Effect of voxel res.}
    \label{fig:voxel}
\end{minipage}
\vspace{-1em}
\end{figure}

{\bf Number of Bins ($N$).} To thoroughly evaluate the effect of different numbers of bins, we conduct a detailed ablation study across varying bin counts as shown in Fig.~\ref{fig:bin}. When $N$ increases from 20 to 80, we observe a consistent decrease in absolute elevation error, which indicates improved elevation discrimination with finer quantization. However, beyond $N = 80$, the performance declines by 3\% to 6\%. This degradation is attributed to over-discretization, where excessive binning results in sparse statistics per bin, increasing sensitivity to noise. So we select $N = 80$ for our final model.

{\bf Voxel Vertical Resolutions.} To investigate the effect of different vertical resolutions of the voxel, we conduct a comprehensive ablation study as shown in Fig.~\ref{fig:voxel}. Overly fine 0.5 cm resolution leads to an absolute error of 2.07 cm, which significantly increases the likelihood of adjacent voxels projecting onto the same image location, complicating the distinction of subtle differences. Moreover, the 1.0 cm resolution emerges as the optimal choice, reducing the absolute error by 10\%. While coarser resolutions ranging from 2.0 to 8.0 cm progressively lose terrain details. At the 8.0 cm resolution, we observe a 3\% increase in absolute error, making it challenging to accurately represent elevation features with rich information.

\begin{figure*}
\centering
\includegraphics[width=\linewidth]{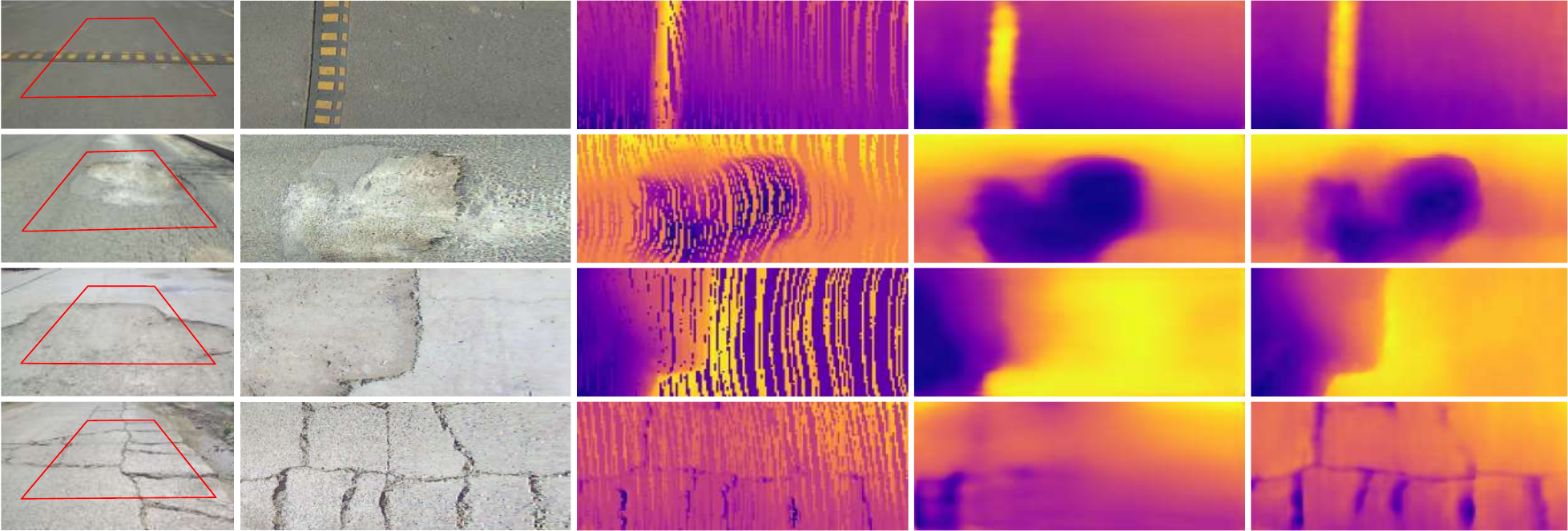}
\caption{Qualitative results on RSRD. From left to right: left images, transformed images in BEV space, ground-truth elevation maps, estimated elevation maps by FastRSR-mono and estimated elevation maps by FastRSR-stereo.}
\label{fig:stereo_vis}
\vspace{-0.5em}
\end{figure*}

{\bf Multi-Scale Projection.} To validate the performance of different multi-scale projection designs, we carefully compare four distinct configurations: Multi-Scale (MS) image projection to MS BEV or Single-Scale (SS) BEV, and various fusion methods of multiple BEV maps including concatenation or addition. As shown in Tab.~\ref{tab:ms}, the MS-to-SS BEV projection outperforms MS-to-MS approaches because the SS BEV representation maintains better spatial consistency, avoiding the feature misalignment and information redundancy. Notably, when employing MS image projection to MS BEV with concatenation, we achieve an optimal trade-off with the absolute elevation error of 1.80 cm and latency of 14.9 ms.
\begin{table}[t]
    \centering
    \caption{Comparison of multi-scale projection designs.}
    \label{tab:ms}
    \small
    \setlength{\tabcolsep}{4.5pt}
    \begin{tabular}{l|l|cccc}
        \hline
        \rowcolor{black!10} \multicolumn{2}{c|}{\textbf{Method}} & \makecell[c]{\textbf{Abs. err.} \\ (\textbf{cm})} & \makecell[c]{\textbf{RMSE} \\ (\textbf{cm})} & \makecell[c]{\textbf{\textgreater0.5 cm} \\ (\textbf{\%})} & \makecell[c]{\textbf{Lantency} \\ (\textbf{ms})}  \\
        \hline
        \multirow{2}{*}{\makecell[l]{MS Image \\ to MS BEV}}
        & plus  & 1.81 & 2.04 & 77.0 & 15.1\\
        & concat  & 1.84 & 2.09 & 77.0 & 15.5\\
        \hline
        \multirow{2}{*}{\makecell[l]{MS Image \\ to SS BEV}}
        & plus  & 1.83 & 2.06 & 76.4 &\textbf{14.6}\\
        & \cellcolor{blue!10}\textbf{concat} & \cellcolor{blue!10}\textbf{1.80} & \cellcolor{blue!10}\textbf{2.01} & \cellcolor{blue!10}\textbf{76.4}&\cellcolor{blue!10}14.9 \\
        \hline
    \end{tabular}
    \vspace{-1em}
\end{table}

\begin{table}[t]
    \centering
    \caption{Comparison of view transformation types.}
    \label{tab:vt} 
    \small
    \setlength{\tabcolsep}{4.0pt}
    \begin{tabular}{c|cccc}
        \hline
        \rowcolor{black!10} \multicolumn{1}{c|} {\textbf{View Transformation}} & \begin{tabular}[c]{@{}c@{}}\textbf{Abs. err.}\\ (cm)\end{tabular}  & \begin{tabular}[c]{@{}c@{}}\textbf{RMSE}\\ (cm)\end{tabular} & \begin{tabular}[c]{@{}c@{}} \textgreater\textbf{0.5 cm}\\ (\%)\end{tabular} & \begin{tabular}[c]{@{}c@{}}\textbf{Latency}\\ (ms)\end{tabular}\\
        \hline
        Bilinear-Sampling~\cite{simplebev}  & 1.80 & 2.02 & 76.4 & 4.21  \\ 
        Look-Up Table~\cite{roadbev} & 1.80 & 2.01 & 76.4 & 0.44 \\
        LSS~\cite{bevdepth} & 1.85 & 2.06 & 76.5 & 0.46 \\
        \rowcolor{blue!10} \textbf{DAP} & \textbf{1.72} & \textbf{1.95}& \textbf{76.3}& \textbf{0.05}\\
        \hline
    \end{tabular}
    \vspace{-1em}
\end{table}

{\bf View Transformation Types.} As shown in Tab.~\ref{tab:vt}, we compare different view transformation operations under the same training settings. When using the look-up table~\cite{roadbev} instead of bilinear-sampling~\cite{simplebev} 3D-to-2D projection, the inference time significantly decreases by 89.5\%. When applying the LSS method~\cite{bevdepth} for 2D-to-3D projection alongside the look-up table, we experience a slight decrease in accuracy with the absolute elevation error of 1.85 cm, which may be attributed to the limitations imposed by the sparse BEV elevation features. The DAP module incorporates depth information and CUDA acceleration operators~\cite{bevpoolv2}, achieving optimal accuracy of 1.72 cm absolute elevation error and the shortest inference time of 0.05 ms, which is over 8 times faster than other methods. Overall, these findings highlight the effective and efficient capabilities of our DAP method in view transformation tasks, demonstrating its potential for enhancing both accuracy and speed.

\subsection{Ablation Studies for FastRSR-stereo}
{\bf Contribution of major components in FastRSR-stereo.} We evaluate the effectiveness of individual components in FastRSR-stereo. We establish our base model on RoadBEV-stereo~\cite{roadbev}. The architecture incorporates a lightweight cost aggregation network and follows a similar paradigm to our previously established monocular BEV feature extraction module, which employs light backbone MobileNetV2~\cite{mobilenetv2} to obtain image features and project image features via the DAP module to derive voxel features. 

As shown in Tab.~\ref{ablation-stereo}, all the modules are beneficial to the performance gains. Using the GC module can lead to a 35.9\% decrease in GPU latency in (a). Subsequently, we employ the SAE module for the adaptive selection of important regions, resulting in 2.8\% absolute elevation error decrease in (b). By incorporating the CAG module for reliable prediction, our model achieves 4.0\% decrease in absolute error in (c). Finally, the complete model reaches the best performance, with an 8.16\% improvement in the absolute elevation error and a 24.6\% decrease in inference time.

\begin{table}[t]
\begin{center}
\caption{Ablation study on the components in FastRSR-stereo. GC represents the group-wise correlation cost volume construction. SAE means using spatial attention enhancement. CAG denotes the confidence attention generation module.}
\label{tab:stereo}
\scalebox{0.78}{ 
  \begin{tabular}{p{1.2cm}p{0.6cm}p{0.8cm}p{0.8cm}|p{1.2cm}p{1.1cm}p{1.1cm}p{1.2cm}}
  
  \hline
 \rowcolor{black!10} \makecell[c]{\textbf{Setting}}&\makecell[c]{\textbf{GC}}& \makecell[c]{\textbf{SAE}}& \makecell[c]{\textbf{CAG}} & \makecell[c]{\textbf{Abs. err.}\\(cm)} &\makecell[c]{\textbf{RMSE}\\(cm)}&\makecell[c]{\textbf{>0.5 cm}\\(\%)}&\makecell[c]{\textbf{Latency}\\(ms)}\\
  \midrule
  
   \makecell[c]{Baseline}&\makecell[c]{\XSolidBrush}& \makecell[c]{\XSolidBrush}& \makecell[c]{\XSolidBrush}& \makecell[c]{0.539} &\makecell[c]{0.662}&\makecell[c]{40.8}&\makecell[c]{28.4}\\
    \hline
\makecell[c]{a}& \makecell[c]{\CheckmarkBold}& \makecell[c]{\XSolidBrush}& \makecell[c]{\XSolidBrush}& \makecell[c]{0.524} &\makecell[c]{0.633}&\makecell[c]{39.7}&\makecell[c]{18.2}\\
 \makecell[c]{b}&\makecell[c]{\CheckmarkBold}& \makecell[c]{\CheckmarkBold}& \makecell[c]{\XSolidBrush}& \makecell[c]{0.509} &\makecell[c]{0.614}&\makecell[c]{38.1}&\makecell[c]{20.1}\\
    \makecell[c]{c}& \makecell[c]{\CheckmarkBold}& \makecell[c]{\XSolidBrush}& \makecell[c]{\CheckmarkBold}& \makecell[c]{0.503} &\makecell[c]{0.613}&\makecell[c]{37.2}&\makecell[c]{19.1}\\
     \hline
   \rowcolor{blue!10}\makecell[c]{d}&\makecell[c]{\CheckmarkBold}& \makecell[c]{\CheckmarkBold}& \makecell[c]{\CheckmarkBold}& \makecell[c]{\textbf{0.495}} &\makecell[c]{\textbf{0.605}}&\makecell[c]{\textbf{36.8}}&\makecell[c]{\textbf{21.4}}\\
  \bottomrule
    \label{ablation-stereo}
  \end{tabular}}
\end{center}
 \vspace{-2.5em}
\end{table}

\begin{table}[t]
    \centering
    \caption{Comparison of different cost volume constructions. }
    \label{tab:cost} 
    \small
    \setlength{\tabcolsep}{4.0pt}
    \begin{tabular}{c|cccc}
        \hline
        \rowcolor{black!10} \multicolumn{1}{c|}{\textbf{Method}} & \begin{tabular}[c]{@{}c@{}}\textbf{Abs. err.}\\ (cm)\end{tabular}  & \begin{tabular}[c]{@{}c@{}}\textbf{RMSE}\\ (cm)\end{tabular} & \begin{tabular}[c]{@{}c@{}} \textgreater\textbf{0.5 cm}\\ (\%)\end{tabular} & \begin{tabular}[c]{@{}c@{}}\textbf{Latency}\\ (ms)\end{tabular}\\  
        \hline
        Diff & 0.539 & 0.662 & 40.8 & 28.4 \\
        Multiply   & \textbf{0.518} &0.635 & 39.7  & 28.5    \\
        Group-Diff   & 0.545 & 0.670 & 40.5  & 18.2 \\
        \rowcolor{blue!10}\textbf{Group-Corr} & 0.524 & \textbf{0.633}& \textbf{39.7}& \textbf{18.2}   \\
        \hline
    \end{tabular}
    \vspace{-1.5em}
\end{table}

{\bf Cost Volume Construction Methods.} To investigate the effect of different cost volume construction methods, we conduct a comprehensive ablation study as shown in Tab.~\ref{tab:cost}. We observe that the inference time is significantly decreased when constructing the group-corr type. This significant speed improvement stems from the parallel group-wise computation. Notably, the group-corr type exhibits only a slight 1\% increase in the absolute elevation error compared to the multiply type. This is negligible compared to the significant 36\% decrease in inference time it brings. The group-corr method demonstrates optimal performance.

\subsection{Qualitative evaluation}

Fig.~\ref{fig:stereo_vis} illustrates the performance of FastRSR models in reconstructing road surface elevation based on monocular or stereo images. In the first scenario with a speed bump, both models effectively capture distinct contours and elevation variations. In the second and third scenarios, which feature significant depressions and potholes, the two models accurately capture the variations in elevation. In the final scenario, where the elevation changes are less pronounced, the FastRSR-stereo model still reveals details of cracks and uneven areas, while the FastRSR-mono model produces a rougher reconstruction. Moreover, the performance is less remarkable at greater distances due to challenges in capturing fine details and potential noise in labels. Overall, this demonstrates the robust reconstruction capabilities of our model across various road surface conditions, confirming its effectiveness for road surface analysis.

\section{CONCLUSIONS}

In this paper, we propose two efficient and accurate models: FastRSR-mono and FastRSR-stereo. To address the limitations of existing view transformation modules for road surface elevation reconstruction, we introduce the Depth-Aware 3D-to-2D Projection (DAP) module, which effectively queries depth and image features to aggregate strong BEV features. To mitigate accuracy loss in the lightweight stereo architectures, we first present the Spatial Attention Enhancement (SAE) module to adaptively select important information. Then, we propose Confidence Attention Generation (CAG) module to highlight high-confidence predictions. Our model demonstrates state-of-the-art performance, enabling efficient and accurate road surface elevation reconstruction.

\bibliographystyle{ACM-Reference-Format}
\bibliography{acmart}


\begin{thebibliography}{56}


\ifx \showCODEN    \undefined \def \showCODEN     #1{\unskip}     \fi
\ifx \showISBNx    \undefined \def \showISBNx     #1{\unskip}     \fi
\ifx \showISBNxiii \undefined \def \showISBNxiii  #1{\unskip}     \fi
\ifx \showISSN     \undefined \def \showISSN      #1{\unskip}     \fi
\ifx \showLCCN     \undefined \def \showLCCN      #1{\unskip}     \fi
\ifx \shownote     \undefined \def \shownote      #1{#1}          \fi
\ifx \showarticletitle \undefined \def \showarticletitle #1{#1}   \fi
\ifx \showURL      \undefined \def \showURL       {\relax}        \fi
\providecommand\bibfield[2]{#2}
\providecommand\bibinfo[2]{#2}
\providecommand\natexlab[1]{#1}
\providecommand\showeprint[2][]{arXiv:#2}

\bibitem[Agarwal and Arora(2023)]%
        {pixelformer}
\bibfield{author}{\bibinfo{person}{Ashutosh Agarwal} {and} \bibinfo{person}{Chetan Arora}.} \bibinfo{year}{2023}\natexlab{}.
\newblock \showarticletitle{Attention attention everywhere: Monocular depth prediction with skip attention}. In \bibinfo{booktitle}{\emph{Proceedings of the IEEE/CVF Winter Conference on Applications of Computer Vision}}. \bibinfo{pages}{5861--5870}.
\newblock


\bibitem[Alrajhi et~al\mbox{.}(2023)]%
        {alrajhi2023detection}
\bibfield{author}{\bibinfo{person}{A Alrajhi}, \bibinfo{person}{K Roy}, \bibinfo{person}{L Qingge}, {and} \bibinfo{person}{J Kribs}.} \bibinfo{year}{2023}\natexlab{}.
\newblock \showarticletitle{Detection of road condition defects using multiple sensors and IoT technology: A review}.
\newblock \bibinfo{journal}{\emph{IEEE Open Journal of Intelligent Transportation Systems}}  \bibinfo{volume}{4} (\bibinfo{year}{2023}), \bibinfo{pages}{372--392}.
\newblock


\bibitem[Beketov and Khalimova(2023)]%
        {beketov2023impact}
\bibfield{author}{\bibinfo{person}{Amir Beketov} {and} \bibinfo{person}{Shakhnoza Khalimova}.} \bibinfo{year}{2023}\natexlab{}.
\newblock \showarticletitle{Impact of roughness and friction properties of road surface of urban streets on the traffic safety}.
\newblock \bibinfo{journal}{\emph{COMMUNICATIONS}} \bibinfo{volume}{25}, \bibinfo{number}{3} (\bibinfo{year}{2023}), \bibinfo{pages}{F51--F63}.
\newblock


\bibitem[Bhat et~al\mbox{.}(2021)]%
        {adabins}
\bibfield{author}{\bibinfo{person}{Shariq~Farooq Bhat}, \bibinfo{person}{Ibraheem Alhashim}, {and} \bibinfo{person}{Peter Wonka}.} \bibinfo{year}{2021}\natexlab{}.
\newblock \showarticletitle{Adabins: Depth estimation using adaptive bins}. In \bibinfo{booktitle}{\emph{Proceedings of the IEEE/CVF conference on computer vision and pattern recognition}}. \bibinfo{pages}{4009--4018}.
\newblock


\bibitem[Blake et~al\mbox{.}(2011)]%
        {markov}
\bibfield{author}{\bibinfo{person}{Andrew Blake}, \bibinfo{person}{Pushmeet Kohli}, {and} \bibinfo{person}{Carsten Rother}.} \bibinfo{year}{2011}\natexlab{}.
\newblock \bibinfo{booktitle}{\emph{Markov random fields for vision and image processing}}.
\newblock \bibinfo{publisher}{MIT press}.
\newblock


\bibitem[Chang and Chen(2018)]%
        {psmnet}
\bibfield{author}{\bibinfo{person}{Jia-Ren Chang} {and} \bibinfo{person}{Yong-Sheng Chen}.} \bibinfo{year}{2018}\natexlab{}.
\newblock \showarticletitle{Pyramid stereo matching network}. In \bibinfo{booktitle}{\emph{Proceedings of the IEEE conference on computer vision and pattern recognition}}. \bibinfo{pages}{5410--5418}.
\newblock


\bibitem[Chen et~al\mbox{.}(2025)]%
        {chen2025stvit+}
\bibfield{author}{\bibinfo{person}{Zhuo Chen}, \bibinfo{person}{Haimei Zhao}, \bibinfo{person}{Xiaoshuai Hao}, \bibinfo{person}{Bo Yuan}, {and} \bibinfo{person}{Xiu Li}.} \bibinfo{year}{2025}\natexlab{}.
\newblock \showarticletitle{STViT+: improving self-supervised multi-camera depth estimation with spatial-temporal context and adversarial geometry regularization}.
\newblock \bibinfo{journal}{\emph{Applied Intelligence}} \bibinfo{volume}{55}, \bibinfo{number}{5} (\bibinfo{year}{2025}), \bibinfo{pages}{328}.
\newblock


\bibitem[Dan et~al\mbox{.}(2024)]%
        {rsrcam3}
\bibfield{author}{\bibinfo{person}{Han-Cheng Dan}, \bibinfo{person}{Bingjie Lu}, {and} \bibinfo{person}{Mengyu Li}.} \bibinfo{year}{2024}\natexlab{}.
\newblock \showarticletitle{Evaluation of asphalt pavement texture using multiview stereo reconstruction based on deep learning}.
\newblock \bibinfo{journal}{\emph{Construction and Building Materials}}  \bibinfo{volume}{412} (\bibinfo{year}{2024}), \bibinfo{pages}{134837}.
\newblock


\bibitem[Duggal et~al\mbox{.}(2019)]%
        {deeppruner}
\bibfield{author}{\bibinfo{person}{Shivam Duggal}, \bibinfo{person}{Shenlong Wang}, \bibinfo{person}{Wei-Chiu Ma}, \bibinfo{person}{Rui Hu}, {and} \bibinfo{person}{Raquel Urtasun}.} \bibinfo{year}{2019}\natexlab{}.
\newblock \showarticletitle{Deeppruner: Learning efficient stereo matching via differentiable patchmatch}. In \bibinfo{booktitle}{\emph{Proceedings of the IEEE/CVF international conference on computer vision}}. \bibinfo{pages}{4384--4393}.
\newblock


\bibitem[Fan et~al\mbox{.}(2018)]%
        {rsrcam1}
\bibfield{author}{\bibinfo{person}{Rui Fan}, \bibinfo{person}{Xiao Ai}, {and} \bibinfo{person}{Naim Dahnoun}.} \bibinfo{year}{2018}\natexlab{}.
\newblock \showarticletitle{Road surface 3D reconstruction based on dense subpixel disparity map estimation}.
\newblock \bibinfo{journal}{\emph{IEEE Transactions on Image Processing}} \bibinfo{volume}{27}, \bibinfo{number}{6} (\bibinfo{year}{2018}), \bibinfo{pages}{3025--3035}.
\newblock


\bibitem[Fan et~al\mbox{.}(2019)]%
        {rsrcam2}
\bibfield{author}{\bibinfo{person}{Rui Fan}, \bibinfo{person}{Jianhao Jiao}, \bibinfo{person}{Jie Pan}, \bibinfo{person}{Huaiyang Huang}, \bibinfo{person}{Shaojie Shen}, {and} \bibinfo{person}{Ming Liu}.} \bibinfo{year}{2019}\natexlab{}.
\newblock \showarticletitle{Real-Time Dense Stereo Embedded in a UAV for Road Inspection}. In \bibinfo{booktitle}{\emph{2019 IEEE/CVF Conference on Computer Vision and Pattern Recognition Workshops (CVPRW)}}. IEEE, \bibinfo{pages}{535--543}.
\newblock


\bibitem[Guo et~al\mbox{.}(2015)]%
        {rsrlidar1}
\bibfield{author}{\bibinfo{person}{Jenny Guo}, \bibinfo{person}{Meng-Ju Tsai}, {and} \bibinfo{person}{Jen-Yu Han}.} \bibinfo{year}{2015}\natexlab{}.
\newblock \showarticletitle{Automatic reconstruction of road surface features by using terrestrial mobile lidar}.
\newblock \bibinfo{journal}{\emph{Automation in Construction}}  \bibinfo{volume}{58} (\bibinfo{year}{2015}), \bibinfo{pages}{165--175}.
\newblock


\bibitem[Guo et~al\mbox{.}(2019)]%
        {gwcnet}
\bibfield{author}{\bibinfo{person}{Xiaoyang Guo}, \bibinfo{person}{Kai Yang}, \bibinfo{person}{Wukui Yang}, \bibinfo{person}{Xiaogang Wang}, {and} \bibinfo{person}{Hongsheng Li}.} \bibinfo{year}{2019}\natexlab{}.
\newblock \showarticletitle{Group-wise correlation stereo network}. In \bibinfo{booktitle}{\emph{Proceedings of the IEEE/CVF conference on computer vision and pattern recognition}}. \bibinfo{pages}{3273--3282}.
\newblock


\bibitem[Guo et~al\mbox{.}(2024)]%
        {lightstereo}
\bibfield{author}{\bibinfo{person}{Xianda Guo}, \bibinfo{person}{Chenming Zhang}, \bibinfo{person}{Dujun Nie}, \bibinfo{person}{Wenzhao Zheng}, \bibinfo{person}{Youmin Zhang}, {and} \bibinfo{person}{Long Chen}.} \bibinfo{year}{2024}\natexlab{}.
\newblock \showarticletitle{LightStereo: Channel Boost Is All Your Need for Efficient 2D Cost Aggregation}.
\newblock \bibinfo{journal}{\emph{arXiv preprint arXiv:2406.19833}} (\bibinfo{year}{2024}).
\newblock


\bibitem[Hao et~al\mbox{.}(2025a)]%
        {HAO2025103018}
\bibfield{author}{\bibinfo{person}{Xiaoshuai Hao}, \bibinfo{person}{Yunfeng Diao}, \bibinfo{person}{Mengchuan Wei}, \bibinfo{person}{Yifan Yang}, \bibinfo{person}{Peng Hao}, \bibinfo{person}{Rong Yin}, \bibinfo{person}{Hui Zhang}, \bibinfo{person}{Weiming Li}, \bibinfo{person}{Shu Zhao}, {and} \bibinfo{person}{Yu Liu}.} \bibinfo{year}{2025}\natexlab{a}.
\newblock \showarticletitle{MapFusion: A novel BEV feature fusion network for multi-modal map construction}.
\newblock \bibinfo{journal}{\emph{Information Fusion}}  \bibinfo{volume}{119} (\bibinfo{year}{2025}), \bibinfo{pages}{103018}.
\newblock


\bibitem[Hao et~al\mbox{.}(2024a)]%
        {hao2024mapdistill}
\bibfield{author}{\bibinfo{person}{Xiaoshuai Hao}, \bibinfo{person}{Ruikai Li}, \bibinfo{person}{Hui Zhang}, \bibinfo{person}{Dingzhe Li}, \bibinfo{person}{Rong Yin}, \bibinfo{person}{Sangil Jung}, \bibinfo{person}{Seung-In Park}, \bibinfo{person}{ByungIn Yoo}, \bibinfo{person}{Haimei Zhao}, {and} \bibinfo{person}{Jing Zhang}.} \bibinfo{year}{2024}\natexlab{a}.
\newblock \showarticletitle{MapDistill: Boosting Efficient Camera-based HD Map Construction via Camera-LiDAR Fusion Model Distillation}.
\newblock \bibinfo{journal}{\emph{arXiv preprint arXiv:2407.11682}} (\bibinfo{year}{2024}).
\newblock


\bibitem[Hao et~al\mbox{.}(2025b)]%
        {hao2025msc}
\bibfield{author}{\bibinfo{person}{Xiaoshuai Hao}, \bibinfo{person}{Guanqun Liu}, \bibinfo{person}{Yuting Zhao}, \bibinfo{person}{Yuheng Ji}, \bibinfo{person}{Mengchuan Wei}, \bibinfo{person}{Haimei Zhao}, \bibinfo{person}{Lingdong Kong}, \bibinfo{person}{Rong Yin}, {and} \bibinfo{person}{Yu Liu}.} \bibinfo{year}{2025}\natexlab{b}.
\newblock \showarticletitle{MSC-Bench: Benchmarking and Analyzing Multi-Sensor Corruption for Driving Perception}.
\newblock \bibinfo{journal}{\emph{arXiv preprint arXiv:2501.01037}} (\bibinfo{year}{2025}).
\newblock


\bibitem[Hao et~al\mbox{.}(2024b)]%
        {hao2024your}
\bibfield{author}{\bibinfo{person}{Xiaoshuai Hao}, \bibinfo{person}{Mengchuan Wei}, \bibinfo{person}{Yifan Yang}, \bibinfo{person}{Haimei Zhao}, \bibinfo{person}{Hui Zhang}, \bibinfo{person}{Yi Zhou}, \bibinfo{person}{Qiang Wang}, \bibinfo{person}{Weiming Li}, \bibinfo{person}{Lingdong Kong}, {and} \bibinfo{person}{Jing Zhang}.} \bibinfo{year}{2024}\natexlab{b}.
\newblock \showarticletitle{Is Your HD Map Constructor Reliable under Sensor Corruptions?}
\newblock \bibinfo{journal}{\emph{arXiv preprint arXiv:2406.12214}} (\bibinfo{year}{2024}).
\newblock


\bibitem[Hao et~al\mbox{.}(2024c)]%
        {hao2024mbfusion}
\bibfield{author}{\bibinfo{person}{Xiaoshuai Hao}, \bibinfo{person}{Hui Zhang}, \bibinfo{person}{Yifan Yang}, \bibinfo{person}{Yi Zhou}, \bibinfo{person}{Sangil Jung}, \bibinfo{person}{Seung-In Park}, {and} \bibinfo{person}{ByungIn Yoo}.} \bibinfo{year}{2024}\natexlab{c}.
\newblock \showarticletitle{Mbfusion: A new multi-modal bev feature fusion method for hd map construction}. In \bibinfo{booktitle}{\emph{2024 IEEE International Conference on Robotics and Automation (ICRA)}}. IEEE, \bibinfo{pages}{15922--15928}.
\newblock


\bibitem[Harley et~al\mbox{.}(2023)]%
        {simplebev}
\bibfield{author}{\bibinfo{person}{Adam~W Harley}, \bibinfo{person}{Zhaoyuan Fang}, \bibinfo{person}{Jie Li}, \bibinfo{person}{Rares Ambrus}, {and} \bibinfo{person}{Katerina Fragkiadaki}.} \bibinfo{year}{2023}\natexlab{}.
\newblock \showarticletitle{Simple-bev: What really matters for multi-sensor bev perception?}. In \bibinfo{booktitle}{\emph{2023 IEEE International Conference on Robotics and Automation (ICRA)}}. IEEE, \bibinfo{pages}{2759--2765}.
\newblock


\bibitem[Huang and Huang(2022)]%
        {bevpoolv2}
\bibfield{author}{\bibinfo{person}{Junjie Huang} {and} \bibinfo{person}{Guan Huang}.} \bibinfo{year}{2022}\natexlab{}.
\newblock \showarticletitle{Bevpoolv2: A cutting-edge implementation of bevdet toward deployment}.
\newblock \bibinfo{journal}{\emph{arXiv preprint arXiv:2211.17111}} (\bibinfo{year}{2022}).
\newblock


\bibitem[Huang et~al\mbox{.}(2021)]%
        {bevdet}
\bibfield{author}{\bibinfo{person}{Junjie Huang}, \bibinfo{person}{Guan Huang}, \bibinfo{person}{Zheng Zhu}, \bibinfo{person}{Yun Ye}, {and} \bibinfo{person}{Dalong Du}.} \bibinfo{year}{2021}\natexlab{}.
\newblock \showarticletitle{Bevdet: High-performance multi-camera 3d object detection in bird-eye-view}.
\newblock \bibinfo{journal}{\emph{arXiv preprint arXiv:2112.11790}} (\bibinfo{year}{2021}).
\newblock


\bibitem[Li et~al\mbox{.}(2024)]%
        {dualbev}
\bibfield{author}{\bibinfo{person}{Peidong Li}, \bibinfo{person}{Wancheng Shen}, \bibinfo{person}{Qihao Huang}, {and} \bibinfo{person}{Dixiao Cui}.} \bibinfo{year}{2024}\natexlab{}.
\newblock \showarticletitle{DualBEV: CNN is All You Need in View Transformation}.
\newblock \bibinfo{journal}{\emph{arXiv preprint arXiv:2403.05402}} (\bibinfo{year}{2024}).
\newblock


\bibitem[Li et~al\mbox{.}(2023a)]%
        {bevstereo}
\bibfield{author}{\bibinfo{person}{Yinhao Li}, \bibinfo{person}{Han Bao}, \bibinfo{person}{Zheng Ge}, \bibinfo{person}{Jinrong Yang}, \bibinfo{person}{Jianjian Sun}, {and} \bibinfo{person}{Zeming Li}.} \bibinfo{year}{2023}\natexlab{a}.
\newblock \showarticletitle{Bevstereo: Enhancing depth estimation in multi-view 3d object detection with temporal stereo}. In \bibinfo{booktitle}{\emph{Proceedings of the AAAI Conference on Artificial Intelligence}}, Vol.~\bibinfo{volume}{37}. \bibinfo{pages}{1486--1494}.
\newblock


\bibitem[Li et~al\mbox{.}(2023b)]%
        {bevdepth}
\bibfield{author}{\bibinfo{person}{Yinhao Li}, \bibinfo{person}{Zheng Ge}, \bibinfo{person}{Guanyi Yu}, \bibinfo{person}{Jinrong Yang}, \bibinfo{person}{Zengran Wang}, \bibinfo{person}{Yukang Shi}, \bibinfo{person}{Jianjian Sun}, {and} \bibinfo{person}{Zeming Li}.} \bibinfo{year}{2023}\natexlab{b}.
\newblock \showarticletitle{Bevdepth: Acquisition of reliable depth for multi-view 3d object detection}. In \bibinfo{booktitle}{\emph{Proceedings of the AAAI Conference on Artificial Intelligence}}, Vol.~\bibinfo{volume}{37}. \bibinfo{pages}{1477--1485}.
\newblock


\bibitem[Li et~al\mbox{.}(2022)]%
        {bevformer}
\bibfield{author}{\bibinfo{person}{Zhiqi Li}, \bibinfo{person}{Wenhai Wang}, \bibinfo{person}{Hongyang Li}, \bibinfo{person}{Enze Xie}, \bibinfo{person}{Chonghao Sima}, \bibinfo{person}{Tong Lu}, \bibinfo{person}{Yu Qiao}, {and} \bibinfo{person}{Jifeng Dai}.} \bibinfo{year}{2022}\natexlab{}.
\newblock \showarticletitle{Bevformer: Learning bird’s-eye-view representation from multi-camera images via spatiotemporal transformers}. In \bibinfo{booktitle}{\emph{European conference on computer vision}}. Springer, \bibinfo{pages}{1--18}.
\newblock


\bibitem[Li et~al\mbox{.}(2023c)]%
        {fbbev}
\bibfield{author}{\bibinfo{person}{Zhiqi Li}, \bibinfo{person}{Zhiding Yu}, \bibinfo{person}{Wenhai Wang}, \bibinfo{person}{Anima Anandkumar}, \bibinfo{person}{Tong Lu}, {and} \bibinfo{person}{Jose~M Alvarez}.} \bibinfo{year}{2023}\natexlab{c}.
\newblock \showarticletitle{Fb-bev: Bev representation from forward-backward view transformations}. In \bibinfo{booktitle}{\emph{Proceedings of the IEEE/CVF International Conference on Computer Vision}}. \bibinfo{pages}{6919--6928}.
\newblock


\bibitem[Liu et~al\mbox{.}(2023)]%
        {bevfusion}
\bibfield{author}{\bibinfo{person}{Zhijian Liu}, \bibinfo{person}{Haotian Tang}, \bibinfo{person}{Alexander Amini}, \bibinfo{person}{Xinyu Yang}, \bibinfo{person}{Huizi Mao}, \bibinfo{person}{Daniela~L Rus}, {and} \bibinfo{person}{Song Han}.} \bibinfo{year}{2023}\natexlab{}.
\newblock \showarticletitle{Bevfusion: Multi-task multi-sensor fusion with unified bird's-eye view representation}. In \bibinfo{booktitle}{\emph{2023 IEEE international conference on robotics and automation (ICRA)}}. IEEE, \bibinfo{pages}{2774--2781}.
\newblock


\bibitem[Mayer et~al\mbox{.}(2016)]%
        {dispnet}
\bibfield{author}{\bibinfo{person}{Nikolaus Mayer}, \bibinfo{person}{Eddy Ilg}, \bibinfo{person}{Philip Hausser}, \bibinfo{person}{Philipp Fischer}, \bibinfo{person}{Daniel Cremers}, \bibinfo{person}{Alexey Dosovitskiy}, {and} \bibinfo{person}{Thomas Brox}.} \bibinfo{year}{2016}\natexlab{}.
\newblock \showarticletitle{A large dataset to train convolutional networks for disparity, optical flow, and scene flow estimation}. In \bibinfo{booktitle}{\emph{Proceedings of the IEEE conference on computer vision and pattern recognition}}. \bibinfo{pages}{4040--4048}.
\newblock


\bibitem[Mei et~al\mbox{.}(2024)]%
        {rome}
\bibfield{author}{\bibinfo{person}{Ruohong Mei}, \bibinfo{person}{Wei Sui}, \bibinfo{person}{Jiaxin Zhang}, \bibinfo{person}{Xue Qin}, \bibinfo{person}{Gang Wang}, \bibinfo{person}{Tao Peng}, \bibinfo{person}{Tao Chen}, {and} \bibinfo{person}{Cong Yang}.} \bibinfo{year}{2024}\natexlab{}.
\newblock \showarticletitle{Rome: Towards large scale road surface reconstruction via mesh representation}.
\newblock \bibinfo{journal}{\emph{IEEE Transactions on Intelligent Vehicles}}  \bibinfo{volume}{9} (\bibinfo{year}{2024}), \bibinfo{pages}{5173--5185}.
\newblock


\bibitem[Michailidis et~al\mbox{.}(2013)]%
        {depth-ele-1}
\bibfield{author}{\bibinfo{person}{Georgios-Tsampikos Michailidis}, \bibinfo{person}{Renato Pajarola}, {and} \bibinfo{person}{Ioannis Andreadis}.} \bibinfo{year}{2013}\natexlab{}.
\newblock \showarticletitle{High performance stereo system for dense 3-D reconstruction}.
\newblock \bibinfo{journal}{\emph{IEEE transactions on circuits and systems for video technology}} \bibinfo{volume}{24}, \bibinfo{number}{6} (\bibinfo{year}{2013}), \bibinfo{pages}{929--941}.
\newblock


\bibitem[Mildenhall et~al\mbox{.}(2021)]%
        {nerf}
\bibfield{author}{\bibinfo{person}{Ben Mildenhall}, \bibinfo{person}{Pratul~P Srinivasan}, \bibinfo{person}{Matthew Tancik}, \bibinfo{person}{Jonathan~T Barron}, \bibinfo{person}{Ravi Ramamoorthi}, {and} \bibinfo{person}{Ren Ng}.} \bibinfo{year}{2021}\natexlab{}.
\newblock \showarticletitle{Nerf: Representing scenes as neural radiance fields for view synthesis}.
\newblock \bibinfo{journal}{\emph{Commun. ACM}} \bibinfo{volume}{65}, \bibinfo{number}{1} (\bibinfo{year}{2021}), \bibinfo{pages}{99--106}.
\newblock


\bibitem[Ni et~al\mbox{.}(2020)]%
        {rsrlidar4}
\bibfield{author}{\bibinfo{person}{Tao Ni}, \bibinfo{person}{Wenhang Li}, \bibinfo{person}{Dingxuan Zhao}, {and} \bibinfo{person}{Zhifei Kong}.} \bibinfo{year}{2020}\natexlab{}.
\newblock \showarticletitle{Road profile estimation using a 3D sensor and intelligent vehicle}.
\newblock \bibinfo{journal}{\emph{Sensors}} \bibinfo{volume}{20}, \bibinfo{number}{13} (\bibinfo{year}{2020}), \bibinfo{pages}{3676}.
\newblock


\bibitem[Oniga and Nedevschi(2009)]%
        {oniga2009processing}
\bibfield{author}{\bibinfo{person}{Florin Oniga} {and} \bibinfo{person}{Sergiu Nedevschi}.} \bibinfo{year}{2009}\natexlab{}.
\newblock \showarticletitle{Processing dense stereo data using elevation maps: Road surface, traffic isle, and obstacle detection}.
\newblock \bibinfo{journal}{\emph{IEEE Transactions on Vehicular Technology}} \bibinfo{volume}{59}, \bibinfo{number}{3} (\bibinfo{year}{2009}), \bibinfo{pages}{1172--1182}.
\newblock


\bibitem[Philion and Fidler(2020)]%
        {lss}
\bibfield{author}{\bibinfo{person}{Jonah Philion} {and} \bibinfo{person}{Sanja Fidler}.} \bibinfo{year}{2020}\natexlab{}.
\newblock \showarticletitle{Lift, splat, shoot: Encoding images from arbitrary camera rigs by implicitly unprojecting to 3d}. In \bibinfo{booktitle}{\emph{Computer Vision--ECCV 2020: 16th European Conference, Glasgow, UK, August 23--28, 2020, Proceedings, Part XIV 16}}. Springer, \bibinfo{pages}{194--210}.
\newblock


\bibitem[Piccinelli et~al\mbox{.}(2023)]%
        {idisc}
\bibfield{author}{\bibinfo{person}{Luigi Piccinelli}, \bibinfo{person}{Christos Sakaridis}, {and} \bibinfo{person}{Fisher Yu}.} \bibinfo{year}{2023}\natexlab{}.
\newblock \showarticletitle{idisc: Internal discretization for monocular depth estimation}. In \bibinfo{booktitle}{\emph{Proceedings of the IEEE/CVF Conference on Computer Vision and Pattern Recognition}}. \bibinfo{pages}{21477--21487}.
\newblock


\bibitem[Roddick et~al\mbox{.}(2019)]%
        {OFTNet}
\bibfield{author}{\bibinfo{person}{Thomas Roddick}, \bibinfo{person}{Alex Kendall}, {and} \bibinfo{person}{Roberto Cipolla}.} \bibinfo{year}{2019}\natexlab{}.
\newblock \showarticletitle{Orthographic Feature Transform for Monocular 3D Object Detection}.
\newblock \bibinfo{journal}{\emph{British Machine Vision Conference}} (\bibinfo{year}{2019}), \bibinfo{pages}{2--3}.
\newblock


\bibitem[Sandler et~al\mbox{.}(2018)]%
        {mobilenetv2}
\bibfield{author}{\bibinfo{person}{Mark Sandler}, \bibinfo{person}{Andrew Howard}, \bibinfo{person}{Menglong Zhu}, \bibinfo{person}{Andrey Zhmoginov}, {and} \bibinfo{person}{Liang-Chieh Chen}.} \bibinfo{year}{2018}\natexlab{}.
\newblock \showarticletitle{Mobilenetv2: Inverted residuals and linear bottlenecks}. In \bibinfo{booktitle}{\emph{Proceedings of the IEEE conference on computer vision and pattern recognition}}. \bibinfo{pages}{4510--4520}.
\newblock


\bibitem[Sch{\"o}nberger et~al\mbox{.}(2016)]%
        {pixelwise}
\bibfield{author}{\bibinfo{person}{Johannes~L Sch{\"o}nberger}, \bibinfo{person}{Enliang Zheng}, \bibinfo{person}{Jan-Michael Frahm}, {and} \bibinfo{person}{Marc Pollefeys}.} \bibinfo{year}{2016}\natexlab{}.
\newblock \showarticletitle{Pixelwise view selection for unstructured multi-view stereo}. In \bibinfo{booktitle}{\emph{Computer Vision--ECCV 2016: 14th European Conference, Amsterdam, The Netherlands, October 11-14, 2016, Proceedings, Part III 14}}. Springer, \bibinfo{pages}{501--518}.
\newblock


\bibitem[Shamsafar et~al\mbox{.}(2022)]%
        {mobilestereonet}
\bibfield{author}{\bibinfo{person}{Faranak Shamsafar}, \bibinfo{person}{Samuel Woerz}, \bibinfo{person}{Rafia Rahim}, {and} \bibinfo{person}{Andreas Zell}.} \bibinfo{year}{2022}\natexlab{}.
\newblock \showarticletitle{Mobilestereonet: Towards lightweight deep networks for stereo matching}. In \bibinfo{booktitle}{\emph{Proceedings of the ieee/cvf winter conference on applications of computer vision}}. \bibinfo{pages}{2417--2426}.
\newblock


\bibitem[Tan and Le(2019)]%
        {efficientnetb0}
\bibfield{author}{\bibinfo{person}{Mingxing Tan} {and} \bibinfo{person}{Quoc~V. Le}.} \bibinfo{year}{2019}\natexlab{}.
\newblock \showarticletitle{EfficientNet: Rethinking Model Scaling for Convolutional Neural Networks}. In \bibinfo{booktitle}{\emph{Proceedings of the International Conference on Machine Learning}}. \bibinfo{publisher}{PMLR}, \bibinfo{address}{Long Beach, CA, USA}, \bibinfo{pages}{6105--6114}.
\newblock


\bibitem[Wang et~al\mbox{.}(2023)]%
        {nerf2}
\bibfield{author}{\bibinfo{person}{Fusang Wang}, \bibinfo{person}{Arnaud Louys}, \bibinfo{person}{Nathan Piasco}, \bibinfo{person}{Moussab Bennehar}, \bibinfo{person}{Luis Rold{\~a}o}, {and} \bibinfo{person}{Dzmitry Tsishkou}.} \bibinfo{year}{2023}\natexlab{}.
\newblock \showarticletitle{Planerf: Svd unsupervised 3d plane regularization for nerf large-scale scene reconstruction}.
\newblock \bibinfo{journal}{\emph{arXiv preprint arXiv:2305.16914}} (\bibinfo{year}{2023}).
\newblock


\bibitem[Wang et~al\mbox{.}(2020b)]%
        {rsrlidar2}
\bibfield{author}{\bibinfo{person}{Lili Wang}, \bibinfo{person}{Dingxuan Zhao}, \bibinfo{person}{Tao Ni}, {and} \bibinfo{person}{Shuang Liu}.} \bibinfo{year}{2020}\natexlab{b}.
\newblock \showarticletitle{Extraction of preview elevation information based on terrain mapping and trajectory prediction in real-time}.
\newblock \bibinfo{journal}{\emph{IEEE Access}}  \bibinfo{volume}{8} (\bibinfo{year}{2020}), \bibinfo{pages}{76618--76631}.
\newblock


\bibitem[Wang et~al\mbox{.}(2020a)]%
        {fadnet}
\bibfield{author}{\bibinfo{person}{Qiang Wang}, \bibinfo{person}{Shaohuai Shi}, \bibinfo{person}{Shizhen Zheng}, \bibinfo{person}{Kaiyong Zhao}, {and} \bibinfo{person}{Xiaowen Chu}.} \bibinfo{year}{2020}\natexlab{a}.
\newblock \showarticletitle{Fadnet: A fast and accurate network for disparity estimation}. In \bibinfo{booktitle}{\emph{2020 IEEE international conference on robotics and automation (ICRA)}}. IEEE, \bibinfo{pages}{101--107}.
\newblock


\bibitem[Woo et~al\mbox{.}(2018)]%
        {cbam}
\bibfield{author}{\bibinfo{person}{Sanghyun Woo}, \bibinfo{person}{Jongchan Park}, \bibinfo{person}{Joon-Young Lee}, {and} \bibinfo{person}{In~So Kweon}.} \bibinfo{year}{2018}\natexlab{}.
\newblock \showarticletitle{Cbam: Convolutional block attention module}. In \bibinfo{booktitle}{\emph{Proceedings of the European conference on computer vision (ECCV)}}. \bibinfo{pages}{3--19}.
\newblock


\bibitem[Xie et~al\mbox{.}(2022)]%
        {M2BEV}
\bibfield{author}{\bibinfo{person}{Enze Xie}, \bibinfo{person}{Zhiding Yu}, \bibinfo{person}{Daquan Zhou}, \bibinfo{person}{Jonah Philion}, \bibinfo{person}{Anima Anandkumar}, \bibinfo{person}{Sanja Fidler}, \bibinfo{person}{Ping Luo}, {and} \bibinfo{person}{Jose~M Alvarez}.} \bibinfo{year}{2022}\natexlab{}.
\newblock \showarticletitle{M2BEV: Multi-Camera Joint 3D Detection and Segmentation with Unified Birds-Eye View Representation}.
\newblock \bibinfo{journal}{\emph{arXiv preprint arXiv:2204.05088}} (\bibinfo{year}{2022}).
\newblock


\bibitem[Xie et~al\mbox{.}(2023)]%
        {nerf1}
\bibfield{author}{\bibinfo{person}{Ziyang Xie}, \bibinfo{person}{Ziqi Pang}, {and} \bibinfo{person}{Yu-Xiong Wang}.} \bibinfo{year}{2023}\natexlab{}.
\newblock \showarticletitle{Mv-map: Offboard hd-map generation with multi-view consistency}. In \bibinfo{booktitle}{\emph{Proceedings of the IEEE/CVF International Conference on Computer Vision}}. \bibinfo{pages}{8658--8668}.
\newblock


\bibitem[Xin et~al\mbox{.}(2024)]%
        {depth-ele-3}
\bibfield{author}{\bibinfo{person}{Yi Xin}, \bibinfo{person}{Siqi Luo}, \bibinfo{person}{Haodi Zhou}, \bibinfo{person}{Junlong Du}, \bibinfo{person}{Xiaohong Liu}, \bibinfo{person}{Yue Fan}, \bibinfo{person}{Qing Li}, {and} \bibinfo{person}{Yuntao Du}.} \bibinfo{year}{2024}\natexlab{}.
\newblock \showarticletitle{Parameter-efficient fine-tuning for pre-trained vision models: A survey}.
\newblock \bibinfo{journal}{\emph{arXiv preprint arXiv:2402.02242}} (\bibinfo{year}{2024}).
\newblock


\bibitem[Xu et~al\mbox{.}(2023b)]%
        {igev}
\bibfield{author}{\bibinfo{person}{Gangwei Xu}, \bibinfo{person}{Xianqi Wang}, \bibinfo{person}{Xiaohuan Ding}, {and} \bibinfo{person}{Xin Yang}.} \bibinfo{year}{2023}\natexlab{b}.
\newblock \showarticletitle{Iterative geometry encoding volume for stereo matching}. In \bibinfo{booktitle}{\emph{Proceedings of the IEEE/CVF Conference on Computer Vision and Pattern Recognition}}. \bibinfo{pages}{21919--21928}.
\newblock


\bibitem[Xu et~al\mbox{.}(2023a)]%
        {acvnet}
\bibfield{author}{\bibinfo{person}{Gangwei Xu}, \bibinfo{person}{Yun Wang}, \bibinfo{person}{Junda Cheng}, \bibinfo{person}{Jinhui Tang}, {and} \bibinfo{person}{Xin Yang}.} \bibinfo{year}{2023}\natexlab{a}.
\newblock \showarticletitle{Accurate and efficient stereo matching via attention concatenation volume}.
\newblock \bibinfo{journal}{\emph{IEEE Transactions on Pattern Analysis and Machine Intelligence}} \bibinfo{volume}{46}, \bibinfo{number}{4} (\bibinfo{year}{2023}), \bibinfo{pages}{2461--2474}.
\newblock


\bibitem[Xu and Zhang(2020)]%
        {aanet}
\bibfield{author}{\bibinfo{person}{Haofei Xu} {and} \bibinfo{person}{Juyong Zhang}.} \bibinfo{year}{2020}\natexlab{}.
\newblock \showarticletitle{Aanet: Adaptive aggregation network for efficient stereo matching}. In \bibinfo{booktitle}{\emph{Proceedings of the IEEE/CVF conference on computer vision and pattern recognition}}. \bibinfo{pages}{1959--1968}.
\newblock


\bibitem[Yang et~al\mbox{.}(2023)]%
        {bevformerv2}
\bibfield{author}{\bibinfo{person}{Chenyu Yang}, \bibinfo{person}{Yuntao Chen}, \bibinfo{person}{Hao Tian}, \bibinfo{person}{Chenxin Tao}, \bibinfo{person}{Xizhou Zhu}, \bibinfo{person}{Zhaoxiang Zhang}, \bibinfo{person}{Gao Huang}, \bibinfo{person}{Hongyang Li}, \bibinfo{person}{Yu Qiao}, \bibinfo{person}{Lewei Lu}, {et~al\mbox{.}}} \bibinfo{year}{2023}\natexlab{}.
\newblock \showarticletitle{Bevformer v2: Adapting modern image backbones to bird's-eye-view recognition via perspective supervision}. In \bibinfo{booktitle}{\emph{Proceedings of the IEEE/CVF Conference on Computer Vision and Pattern Recognition}}. \bibinfo{pages}{17830--17839}.
\newblock


\bibitem[Zhao et~al\mbox{.}(2023)]%
        {rsrlidar3}
\bibfield{author}{\bibinfo{person}{Tong Zhao}, \bibinfo{person}{Peilin Guo}, \bibinfo{person}{Junxiang He}, {and} \bibinfo{person}{Yintao Wei}.} \bibinfo{year}{2023}\natexlab{}.
\newblock \showarticletitle{A hierarchical scheme of road unevenness perception with lidar for autonomous driving comfort}.
\newblock \bibinfo{journal}{\emph{IEEE Transactions on Intelligent Vehicles}} \bibinfo{volume}{9}, \bibinfo{number}{1} (\bibinfo{year}{2023}), \bibinfo{pages}{2439--2448}.
\newblock


\bibitem[Zhao et~al\mbox{.}(2024a)]%
        {rsrd}
\bibfield{author}{\bibinfo{person}{Tong Zhao}, \bibinfo{person}{Yichen Xie}, \bibinfo{person}{Mingyu Ding}, \bibinfo{person}{Lei Yang}, \bibinfo{person}{Masayoshi Tomizuka}, {and} \bibinfo{person}{Yintao Wei}.} \bibinfo{year}{2024}\natexlab{a}.
\newblock \showarticletitle{A road surface reconstruction dataset for autonomous driving}.
\newblock \bibinfo{journal}{\emph{Scientific data}} \bibinfo{volume}{11}, \bibinfo{number}{1} (\bibinfo{year}{2024}), \bibinfo{pages}{459}.
\newblock


\bibitem[Zhao et~al\mbox{.}(2024b)]%
        {roadbev}
\bibfield{author}{\bibinfo{person}{Tong Zhao}, \bibinfo{person}{Lei Yang}, \bibinfo{person}{Yichen Xie}, \bibinfo{person}{Mingyu Ding}, \bibinfo{person}{Masayoshi Tomizuka}, {and} \bibinfo{person}{Yintao Wei}.} \bibinfo{year}{2024}\natexlab{b}.
\newblock \showarticletitle{RoadBEV: Road Surface Reconstruction in Bird's Eye View}.
\newblock \bibinfo{journal}{\emph{arXiv preprint arXiv:2404.06605}} (\bibinfo{year}{2024}).
\newblock


\bibitem[Zhou et~al\mbox{.}(2023)]%
        {matrixvt}
\bibfield{author}{\bibinfo{person}{Hongyu Zhou}, \bibinfo{person}{Zheng Ge}, \bibinfo{person}{Zeming Li}, {and} \bibinfo{person}{Xiangyu Zhang}.} \bibinfo{year}{2023}\natexlab{}.
\newblock \showarticletitle{Matrixvt: Efficient multi-camera to bev transformation for 3d perception}. In \bibinfo{booktitle}{\emph{Proceedings of the IEEE/CVF International Conference on Computer Vision}}. \bibinfo{pages}{8548--8557}.
\newblock


\end{thebibliography}

\end{document}